\documentclass[conference]{IEEEtran}
\IEEEoverridecommandlockouts

\usepackage{cite}
\usepackage[table,dvipsnames]{xcolor}
\usepackage{amsmath,amssymb,amsfonts}
\usepackage{algorithmic}
\usepackage{graphicx}
\usepackage{textcomp}
\usepackage{booktabs}
\usepackage{multirow}
\usepackage{caption}
\usepackage{comment}
\usepackage{pifont}
\usepackage[caption=false,font=footnotesize]{subfig}
\usepackage{tabularx}
\usepackage{url}
\usepackage{placeins}
\usepackage{stfloats}
\usepackage{float}
\usepackage[hidelinks]{hyperref}

\newcommand{\cmark}{\textcolor{green!60!black}{\ding{51}}}
\newcommand{\xmark}{\textcolor{red}{\ding{55}}}

\title{Do Vision-Language-Action Foundation Models Generalize to Mobile Robot Navigation?}

\author{\IEEEauthorblockN{Sahar Salimpour and Murtaza Hazara}
\IEEEauthorblockA{Nokia Bell Labs\\
Email: sahar.salimpour@nokia.com, murtaza.hazara@nokia-bell-labs.com}}

\begin{document}

\maketitle

\begin{abstract}
Vision--language--action (VLA) models are emerging as robot foundation policies, yet their generalization beyond manipulation remains insufficiently understood. Mobile robot navigation provides a complementary test because policies must map visual observations, proprioceptive state, and language instructions to closed-loop linear and angular velocity commands under changes in geometry, scene, target semantics, command order, and sensing modality. We introduce a unified benchmark for VLA-based mobile navigation with five task categories: primitive motion control, trajectory extension, compositional navigation, semantic goal grounding, and safety-constrained navigation with modality dropout. The benchmark defines controlled in-distribution (ID) and out-of-distribution (OOD) evaluations over spatial, visual, semantic, procedural, and sensor-modality shifts. We fine-tune and compare GR00T N1.6, pi0.5, OpenVLA-OFT, SmolVLA, and ACT as end-to-end navigation policies, and validate a selected policy on a physical mobile robot. Results show that GR00T achieves the strongest ID performance and the most reliable closed-loop navigation, while pi0.5 is more robust to some procedural command shifts. Across models, visual shifts such as new scenes or camera views are generally less disruptive than semantic target changes, ordered-command changes, and robot-state shifts. Modality-dropout in obstacle-navigation experiments indicate different sensing biases: GR00T depends more on proprioceptive state, whereas pi0.5 is more sensitive to visual inputs. These findings show that current VLAs can be adapted to mobile navigation, but robust generalization remains limited by state/action grounding, semantic grounding, procedural composition, and sensor dependence.
\end{abstract}


\section{Introduction}
Foundation models that couple visual and language perception to generate actions, are reshaping robotics by narrowing the gap between general-purpose AI and embodied decision making. Recent generalist vision--language--action (VLA) models are pretrained on large, heterogeneous robot datasets spanning multiple embodiments, environments, and tasks, enabling promising few-shot and ideally zero-shot transfer to new settings~\cite{bjorck2025gr00t, kim2024openvla, intelligence2025pi, zha2026lap}. Recent work on cross-embodiment generalization has largely focused on manipulation-centric tasks~\cite{li2026pointvla, he2025capsdt}, where high-quality open-source datasets are readily available~\cite{o2024open, bu2025agibot_iros,liu2023libero}, while mobile robot navigation, a distinct task family with different embodiment, action space, and sensing modalities, remains relatively underexplored. 
\begin{figure}[!t]
\centering
\includegraphics[width=\columnwidth, height=0.8\columnwidth]{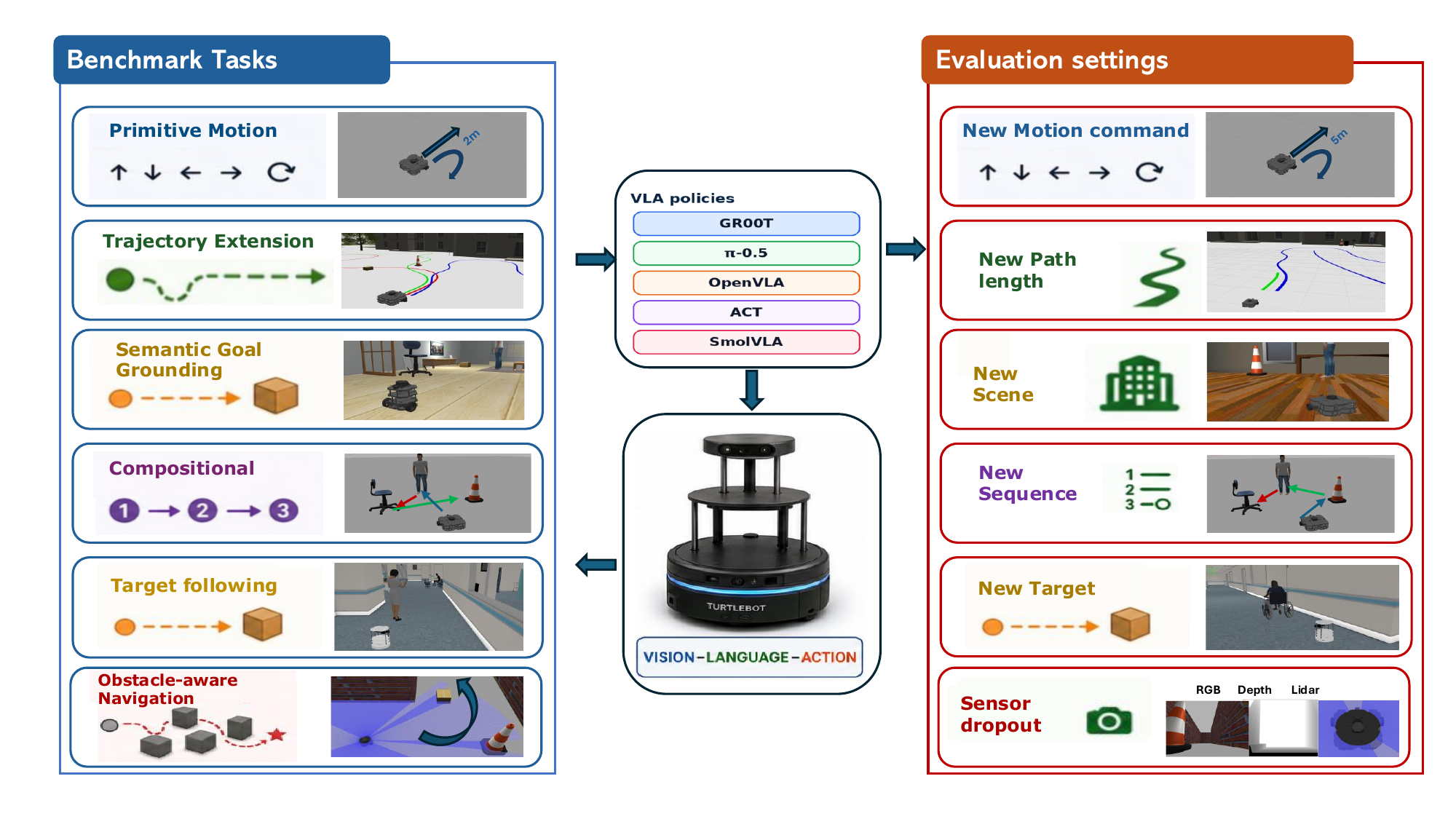}
\captionsetup{justification=raggedright,singlelinecheck=false}
\caption{Overview of the proposed VLA mobile-navigation benchmark, including example task categories, and OOD generalization shifts.}
\label{fig:combined}
\end{figure}

\textbf{Evaluating and quantifying generalization.}
Progress toward truly generalist VLAs requires evaluation protocols that offer diversity of tasks and shifts, controlled coverage of deployment-relevant generalization factors, and reproducible evaluation procedures.
Despite rapid progress, rigorous and challenge-driven benchmarks remain scarce. Existing works often evaluate primarily manipulation tasks~\cite{fei2025libero, wang2025vlatest}, using open-source datasets and focusing mainly on robustness to visual and language distractors~\cite{gao2026taxonomy}. While these efforts provide useful baselines for comparing models under standardized metrics, systematic evaluation under deployment-relevant distribution shifts remains limited. Furthermore, end-to-end deployment on novel embodiments and task families, such as mobile robot navigation, where open-source task-specific datasets are sparse or unavailable, remains overlooked.

\textbf{Motivation and Objectives.}
To address these limitations, we evaluate state-of-the-art VLA models on mobile robot navigation, with targeted, task specific distribution-shift testing. Compared to many manipulation setups, mobile navigation offers a straightforward deployment pipeline, easy reproducibility across simulation and real-world environments, and a modular task structure that enables controlled evaluation of distinct generalization factors. We introduce a unified benchmark comprising various task categories, each designed to probe a specific generalization capability, enabling systematic, and fine-grained comparison across in-distribution (ID) and out-of-distribution (OOD) conditions. Our objective is to quantify OOD robustness across representative VLA models and characterize each model's strengths and weaknesses in terms of generalization across vision, language, and state/action spaces.

\textbf{Contribution.}
The main contributions of this paper are as follows:

\textbf{-} We introduce multiple navigation scenarios across five diverse task categories: \textit{primitive motion control}, \textit{trajectory extension}, \textit{compositional tasks}, \textit{semantic goal grounding}, and \textit{safety constraint modality dropout} based on a unified evaluation protocol.

\textbf{-} For each task, we define distinct OOD evaluation settings to probe complementary dimensions of generalization over vision, language, and state/action spaces. We categorize these perturbations into \textit{spatial}, \textit{visual}, \textit{semantic}, \textit{procedural}, and \textit{modality dropout} groups to provide a human-interpretable framework for policy evaluation.

\textbf{-} We fine-tune and evaluate representative state-of-the-art VLA models, \textbf{GR00T N1.6}, \textbf{\boldmath$\pi$0.5}, \textbf{SmolVLA}, \textbf{ACT}, and \textbf{OpenVLA-OFT} on this benchmark, providing a qualitative and quantitative comparison of their generalization behavior.

\section{Related Work}
\label{sec:related}

\textbf{VLA Foundation Models}
Recent advances in VLA models have been enabled by large-scale, multi-embodiment datasets that aggregate robot data across diverse tasks and platforms\cite{o2024open}. Building on this foundation, several notable VLA architectures have emerged. OpenVLA \cite{kim2024openvla} provides the first fully open-source VLA implementation for manipulation tasks. GR00T \cite{bjorck2025gr00t} extends VLA capabilities to humanoid robots, demonstrating whole-body control for high-DOF systems. The $\pi_0$ model \cite{intelligence2025pi} emphasizes zero-shot generalization of manipulation policies to novel objects and scenes beyond the training distribution. Domain-specific adaptations include MoManipVLA\cite{wu2025momanipvla} for mobile manipulation tasks such as pick-and-place with base locomotion, CapsDT \cite{he2025capsdt} for medical robotics applications, and SmolVLA \cite{shukor2025smolvla}, which achieves competitive performance with significantly reduced computational requirements. These VLA models typically combine pretrained vision and language encoders with action expert modules to directly output low-level control commands. In contrast, ACT~\cite{zhao2023learning} is not a VLA foundation model but a vision-based imitation-learning policy that uses transformer action chunking to reduce compounding errors in fine-grained manipulation.

\textbf{VLA Models for Robotic Navigation}
While VLA models have primarily targeted manipulation, recent work has begun
exploring their application to mobile robot navigation~\cite{shah2023lm}.
MM-Nav~\cite{xu2025mm} trains a multiview (360$^{\circ}$) VLA navigation
policy in a teacher--student manner, distilling behaviors from multiple RL
experts. OmniVLA~\cite{hirose2025omnivla} proposes an omni-modal goal-conditioned
VLA navigation framework, built on top of OpenVLA, trained to follow goals
specified via multiple complementary modalities. WMNav~\cite{nie2025wmnav}
introduces a VLM-powered world-model navigation framework for zero-shot object
goal navigation with a two-stage action proposer, it mitigates hallucination and
improves success and exploration efficiency in unseen environments. Most VLA
approches for navigation remain primarily VLM-powered, which even called
vision-language-navigation(VLN) models, with their action space and downstream
components tailored specifically for navigation tasks. As such, they differ in
scope from the generalist VLA models considered in this work.

\textbf{Benchmarking and Evaluation of VLA Policies}
Rigorous evaluation of VLA models has become increasingly critical as the field matures. The LIBERO benchmark~\cite{liu2023libero} provides standardized task suites for evaluating transfer learning and continual learning in manipulation. Zhang et al.~\cite{zhang2025experiences} benchmark representative VLAs with unified metrics for efficiency, OOD adaptability, and instruction-following, and distill practical failure modes and deployment trade-offs. Fang et al.~\cite{fang2025intention,fang2025intention} propose INT-ACT, a simulation probing benchmark for VLA generalization, and identify an intention--action gap where strong semantic understanding under OOD shift does not reliably translate to correct execution. VLA-Test~\cite{wang2025vlatest} introduces automated testing methodologies adapted from software engineering to ensure systematic validation of VLA policies. VLA-Test~\cite{wang2025vlatest} introduces a fuzzing-style scene generation framework for systematically testing VLA robustness under visual and language perturbations. Despite these advances, existing benchmarks predominantly focus on manipulation tasks, leaving mobile navigation as a relatively underexplored evaluation domain for VLA models.

\section{A Benchmark for VLA-Based Mobile Navigation}
\label{sec:benchmark}

\subsection{Data Collection}
\label{sec:data_collection}
We develop and validate our data collection and evaluation pipeline in simulation using the Gazebo simulator with ROS2 Humble to support reproducibility and future comparisons. Across the benchmark, the core observations consist of egocentric RGB images and robot odometry, with state data represented relative to the robot odometry reference frame to avoid reliance on absolute position. Specific tasks additionally use task-relevant sensing modalities; these variations are detailed alongside the task categories in Section~\ref{sec:tasks_category_design}. The policies produce linear \(x\) and angular \(z\) velocity commands. For real-world validation, we deploy on a Kelo Robile mobile base and use an Intel RealSense D435if as the RGB camera.

\begin{table}[h!]
\centering
\caption{Model fine-tuning details used in the benchmark.}
\label{tab:training_details}
\scriptsize
\resizebox{\columnwidth}{!}{%
\begin{tabular}{lcccc}
\toprule
\textbf{Model} & \textbf{Base Model} & \textbf{Steps} & \textbf{Batch Size} & \textbf{Compute Budget} \\
\midrule
GR00T N1.6 & NVIDIA GR00T-N1.6-3B & 10k & 32 & 1 A100 80GB \\
\midrule
\(\pi\)0.5 & \(\pi\)0.5 & 30k & 32 & 1 A100 80GB \\
\midrule
OpenVLA-OFT & OpenVLA-7B & 20k & 8 & 1 A100 80GB \\
\midrule
SmolVLA & SmolVLA base & 50k & 32 & 1 A100 80GB \\
\midrule
ACT & ACT & 50k & 32 & 1 A100 80GB \\
\bottomrule
\end{tabular}%
}
\end{table}

\begin{table*}[t]
\centering
\caption{Navigation Tasks for VLA Benchmarking.}
\tiny
\label{tab:tasks_categories}
\renewcommand{\arraystretch}{1.18}
\setlength{\tabcolsep}{1pt}
\resizebox{\textwidth}{!}{%
\begin{tabular}{p{1.5cm}p{1.8cm}p{3.0cm}p{7.9cm}}
\toprule
\textbf{Category} & \textbf{Tasks} & \textbf{Example Command} & \multicolumn{1}{c}{\textbf{Example Figures}} \\
\midrule

\multirow{3}{*}{\begin{tabular}[c]{@{}l@{}}{\textbf{C1.}} \\ {\textbf{Primitive}}\\ {\textbf{Motion Control}}\end{tabular}}
& {\fontsize{4.8}{5.2}\selectfont \textbf{C11. Translation}} & {\fontsize{4.5}{5}\selectfont ``Move N meters forward''} & \multirow{3}{*}{\includegraphics[height=1.4cm]{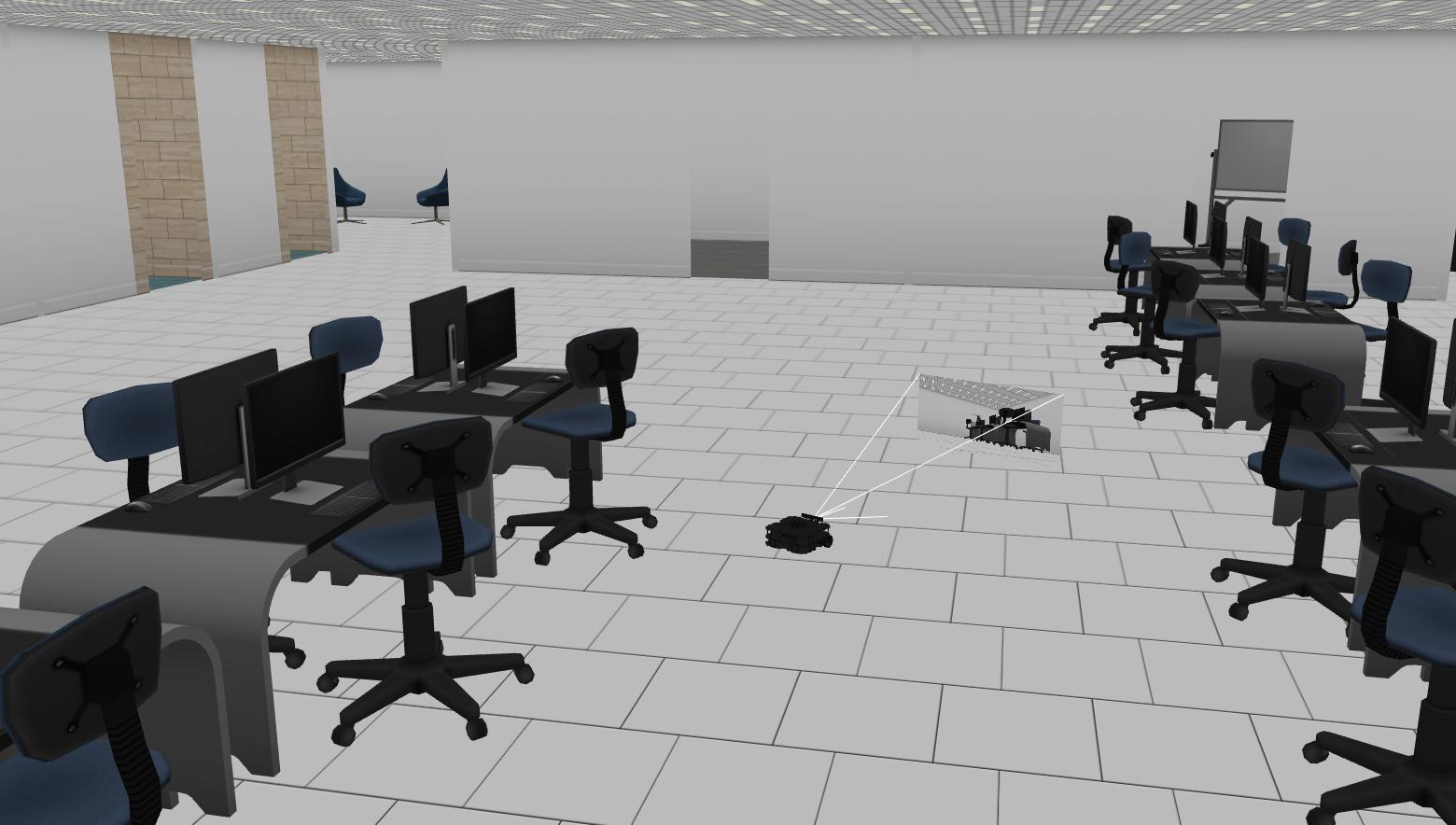}\hspace{1.5mm}\includegraphics[height=1.4cm]{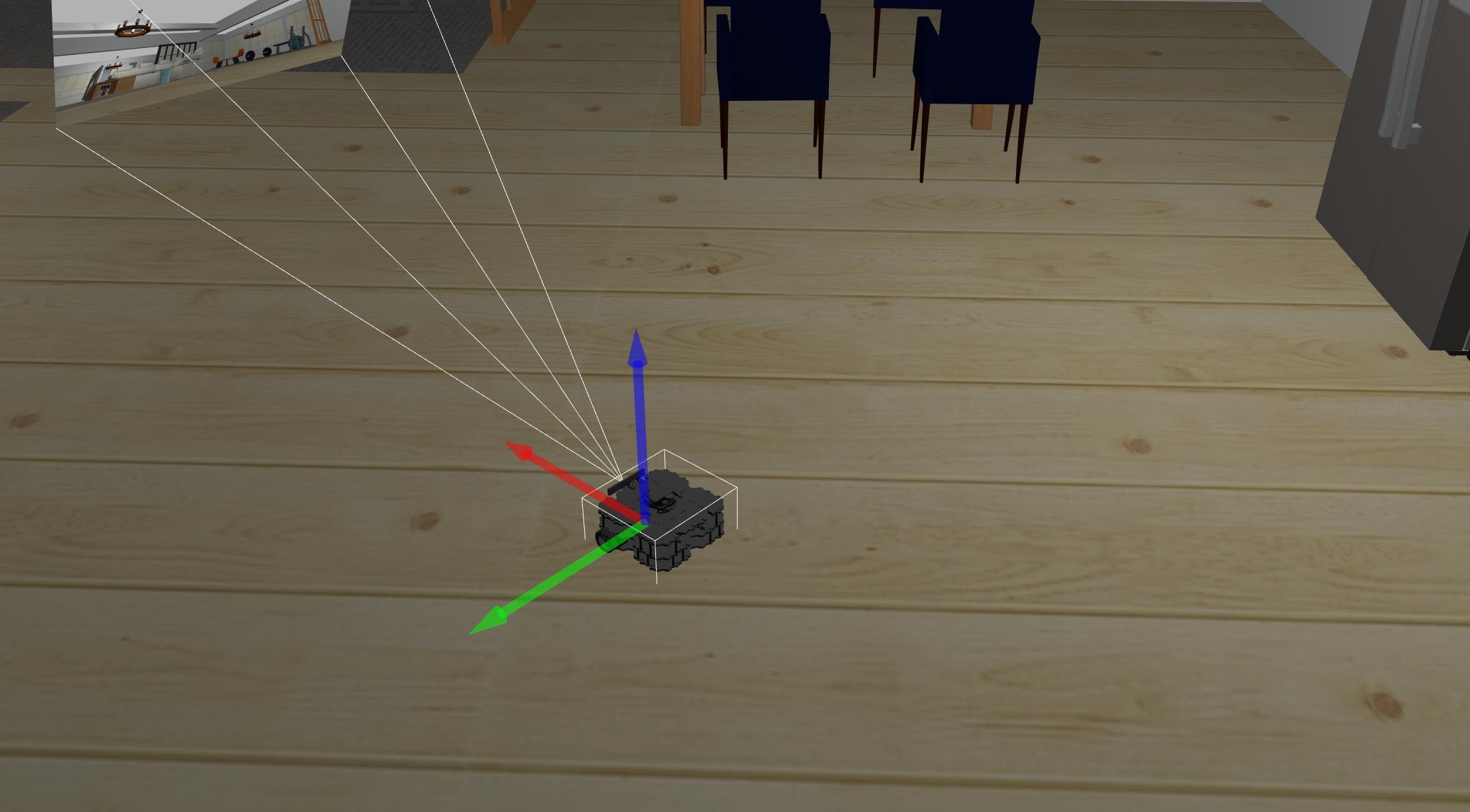}\hspace{1.5mm}\includegraphics[height=1.4cm]{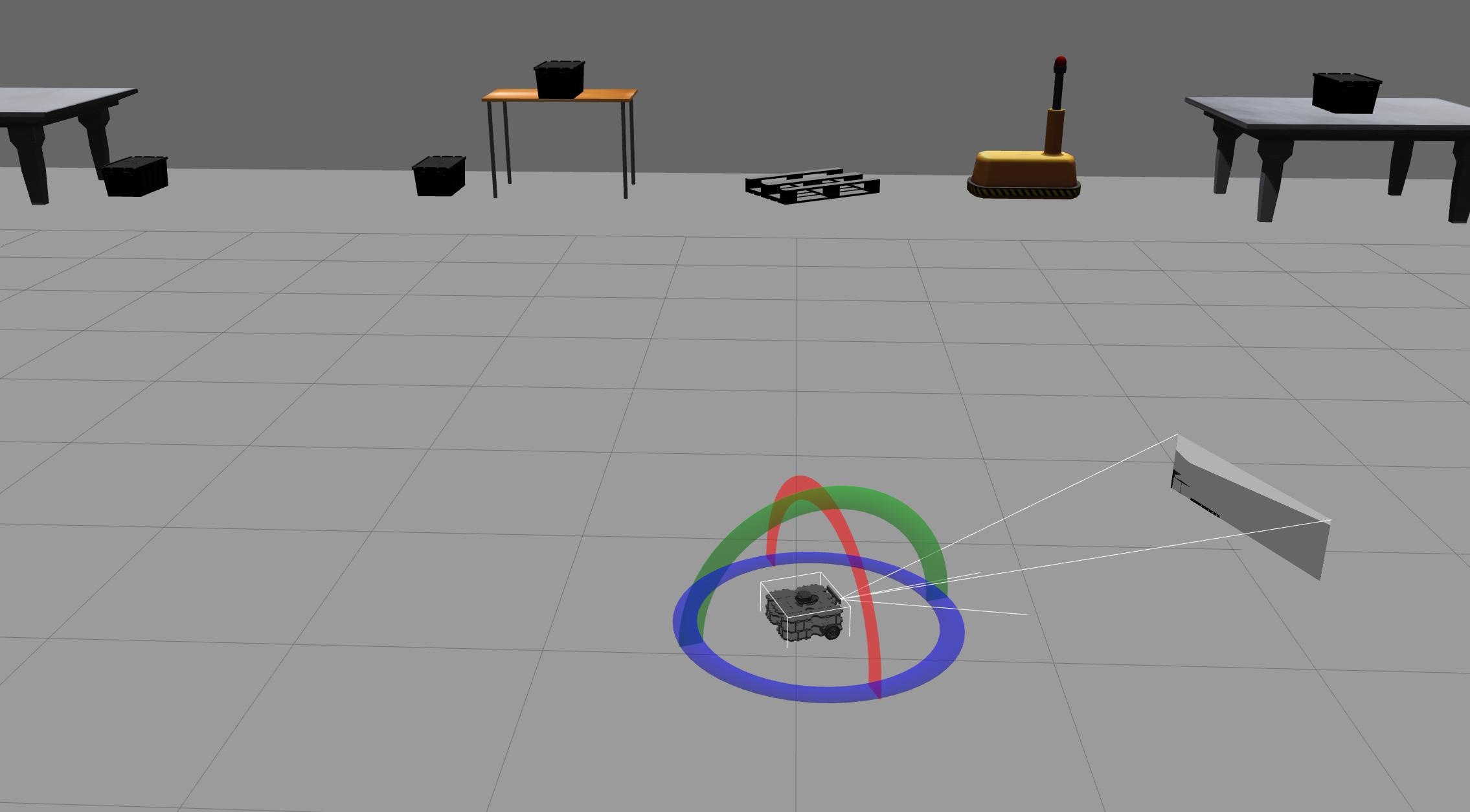}} \\
\cmidrule(lr){2-3}
& {\fontsize{4.8}{5.2}\selectfont \textbf{C12. Rotation}} & {\fontsize{4.5}{5}\selectfont ``Rotate N degrees''} & \\
\cmidrule(lr){2-3}
& \begin{tabular}[c]{@{}l@{}}{\fontsize{4.8}{5.2}\selectfont \textbf{C13. Short action}} \\ {\fontsize{4.8}{5.2}\selectfont \textbf{composition}}\end{tabular} & \begin{tabular}[c]{@{}l@{}}{\fontsize{4.5}{5}\selectfont ``Rotate N degrees and then } \\ {\fontsize{4.5}{5}\selectfont move N meters backward''}\end{tabular} & \\
\addlinespace[5pt]
\midrule

\raisebox{-0.19cm}{\begin{tabular}[c]{@{}l@{}}{\textbf{C2.}} \\ {\textbf{Trajectory}} \\ {\textbf{Extension}}\end{tabular}}
& {\fontsize{4.8}{5.2}\selectfont \textbf{C21. Line following}} & \begin{tabular}[c]{@{}l@{}}{\fontsize{4.5}{5}\selectfont ``Follow the red line''}\end{tabular} & \multirow{2}{*}{\includegraphics[height=1.4cm]{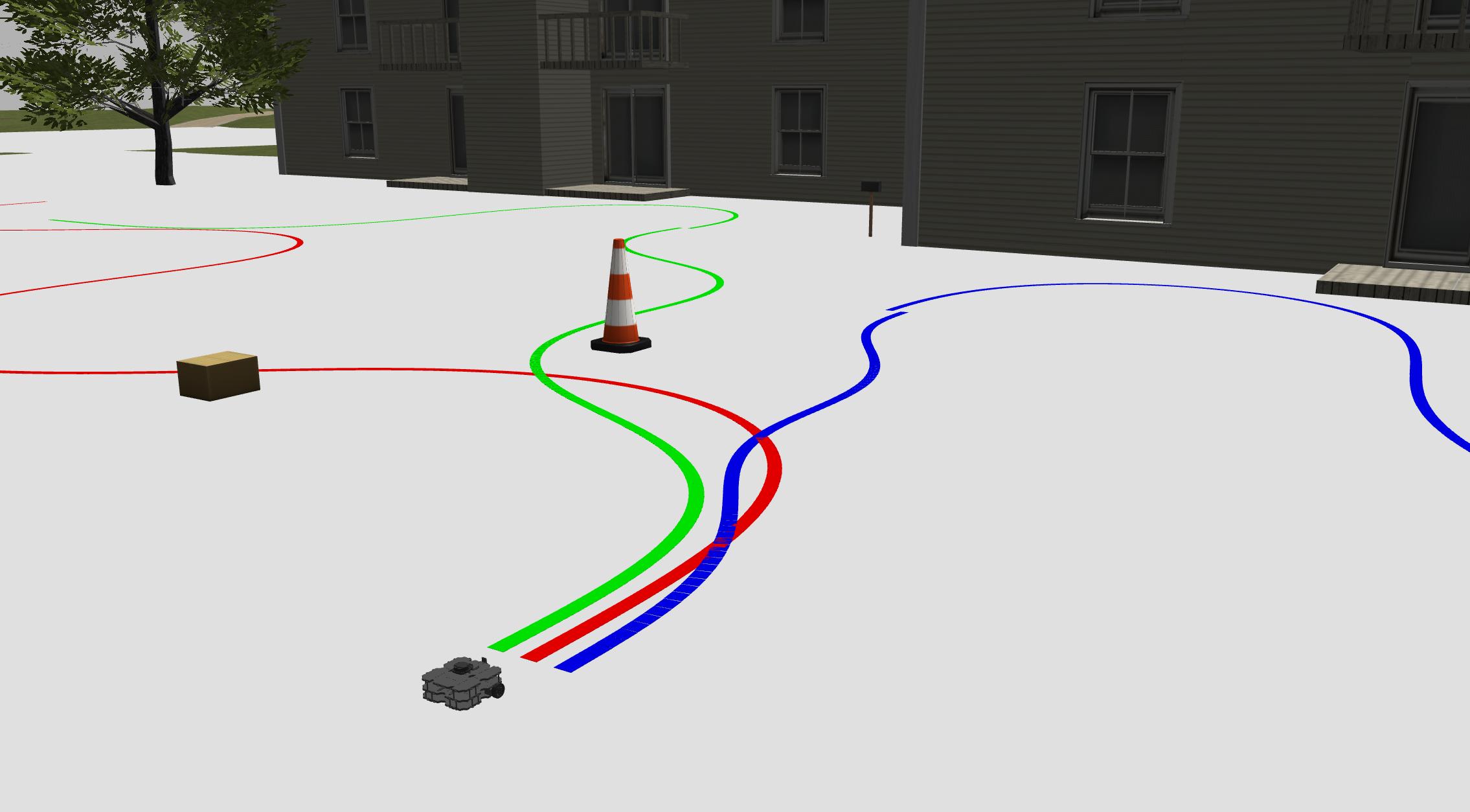}\hspace{1.6mm}\includegraphics[height=1.4cm]{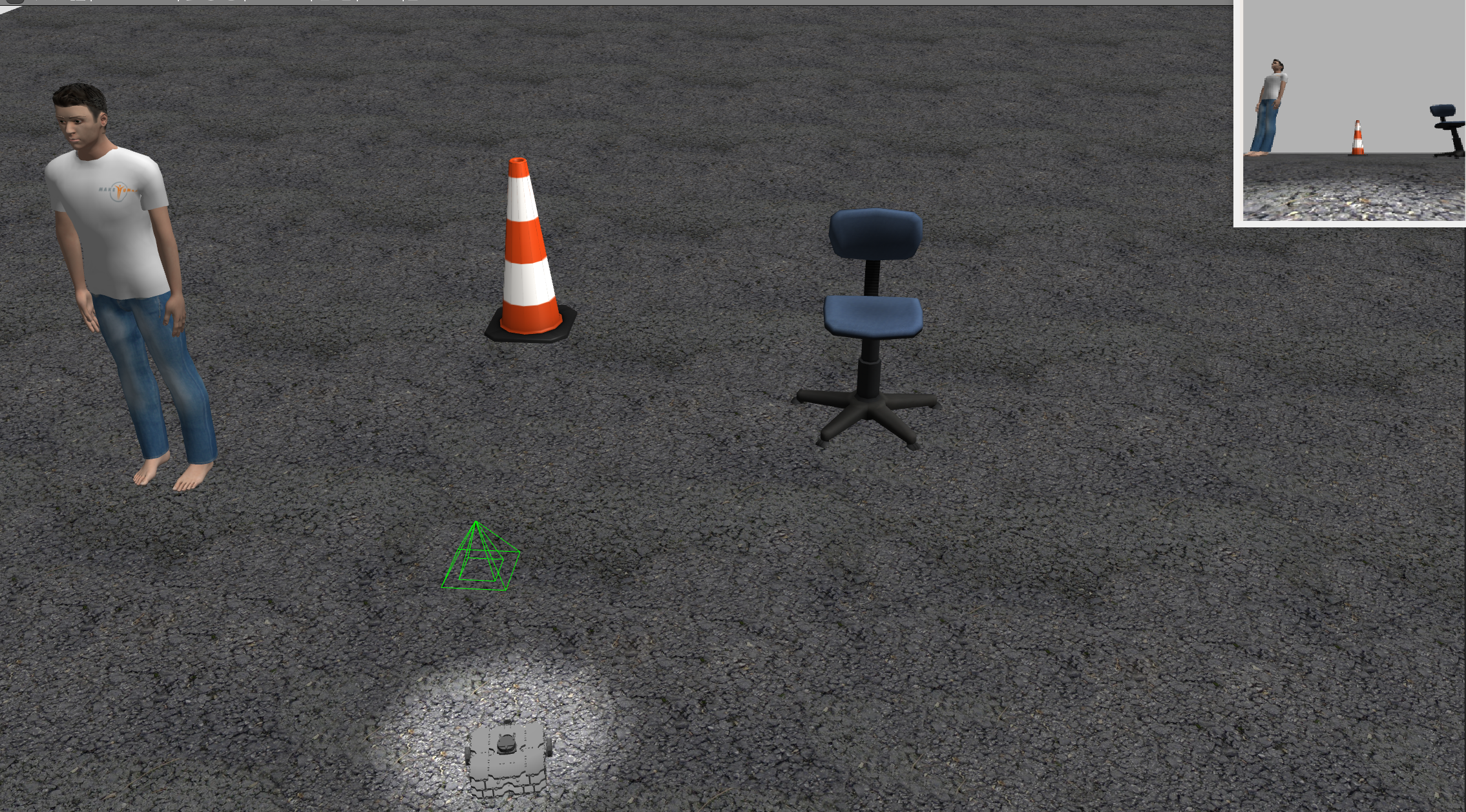}\hspace{2.0mm}\includegraphics[height=1.4cm]{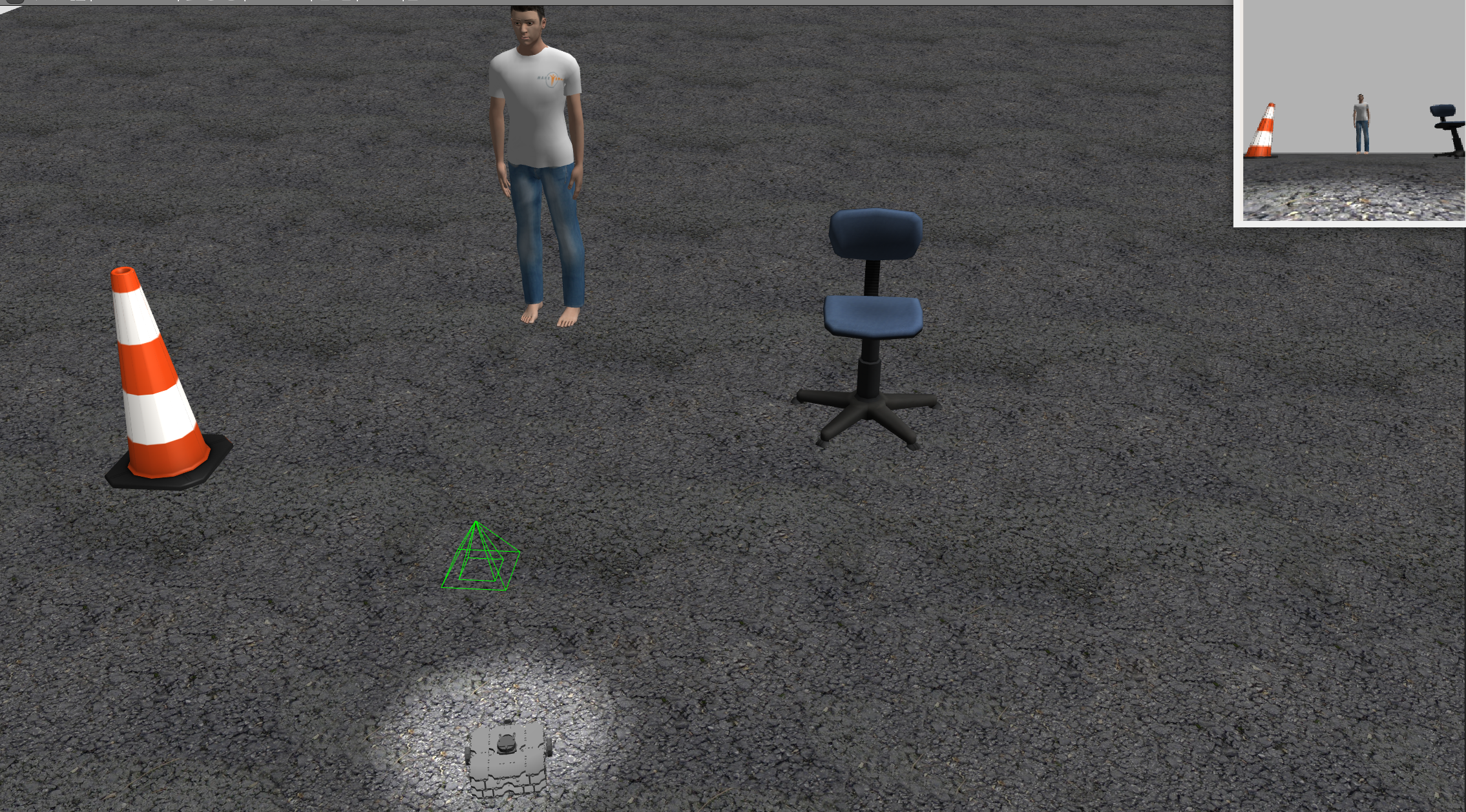}} \\
\cmidrule[\lightrulewidth](r){1-3}
\begin{tabular}[c]{@{}l@{}}{\textbf{C3.}} \\ {\textbf{Compositional}} \\ {\textbf{Tasks}}\end{tabular}
& {\fontsize{4.8}{5.2}\selectfont \textbf{C31. Targets sequence}} & \begin{tabular}[c]{@{}l@{}}{\fontsize{4.5}{5}\selectfont ``navigate to the person, then cone,} \\ {\fontsize{4.5}{5}\selectfont and then the chair''}\end{tabular} &  \\
\addlinespace[-2pt]
\midrule

\multirow{3}{*}{\begin{tabular}[c]{@{}l@{}}{\textbf{C4. }} \\ {\textbf{Semantic}} \\ {\textbf{Goal Grounding}}\end{tabular}}
& {\fontsize{4.8}{5.2}\selectfont \textbf{C41. Goal-directed navigation}} & {\fontsize{4.5}{5}\selectfont ``Navigate to the TV / cone''} & \multirow{3}{*}{\includegraphics[height=1.4cm]{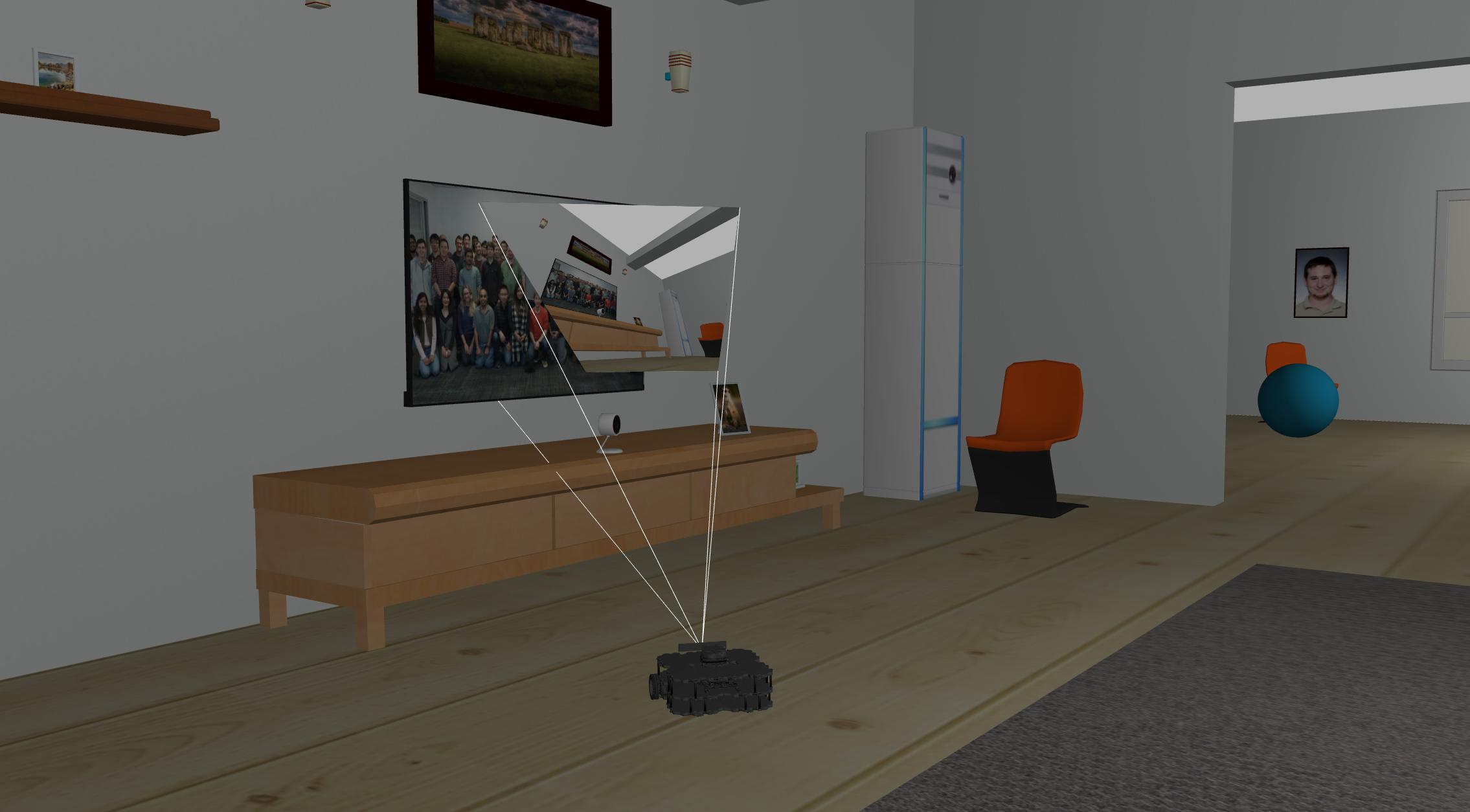}\hspace{1.5mm}\includegraphics[height=1.4cm]{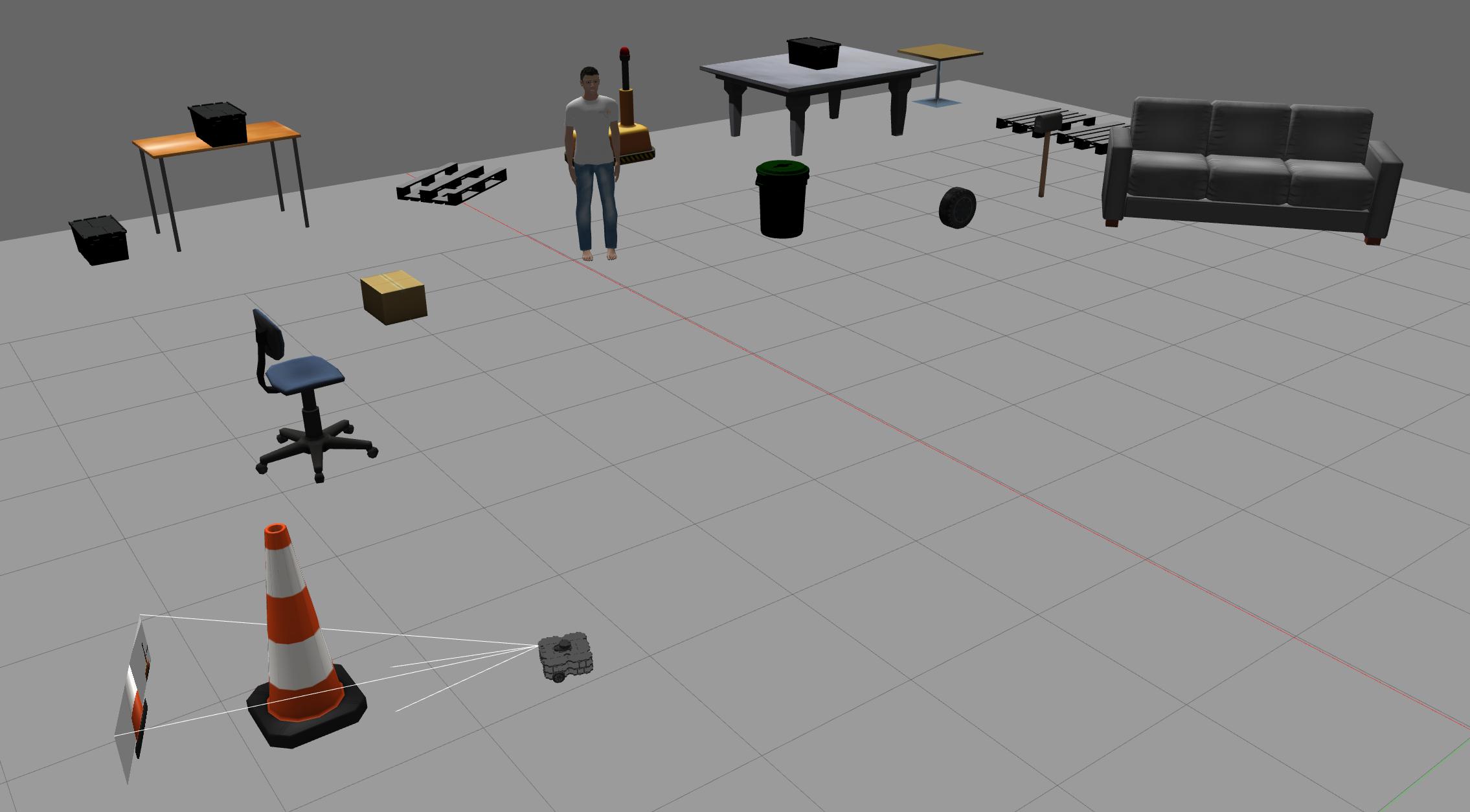}\hspace{1.5mm}\includegraphics[height=1.4cm]{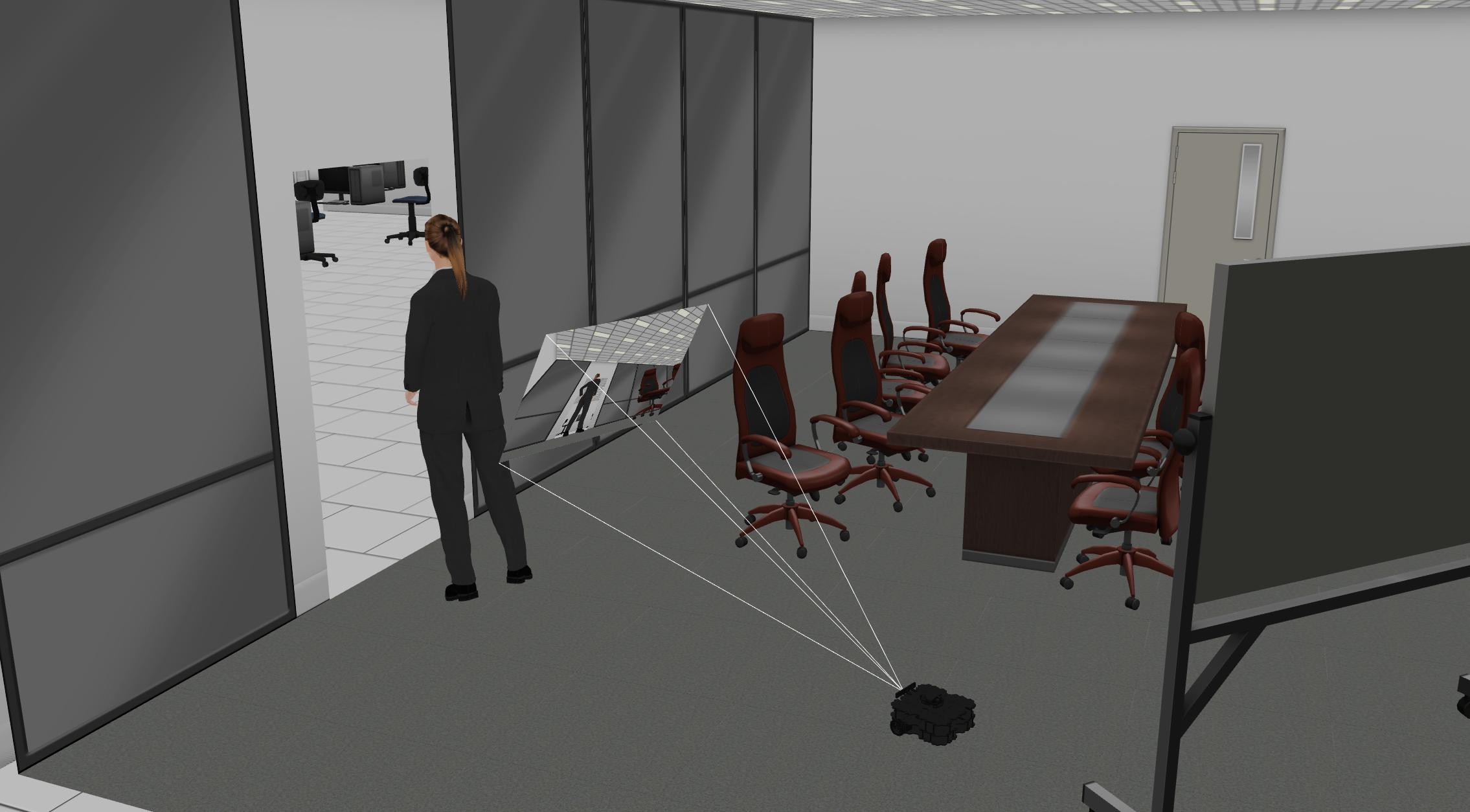}} \\
\cmidrule(lr){2-3}
& {\fontsize{4.8}{5.2}\selectfont \textbf{C42. Dynamic Target following}} & {\fontsize{4.5}{5}\selectfont ``Follow the person''} & \\
\addlinespace[12pt]
\bottomrule

\multirow{3}{*}{\begin{tabular}[c]{@{}l@{}}{\textbf{C5.}} \\ {\textbf{Safety}}\\ {\textbf{Constraint}} \\ {\textbf{Modality}} \\ {\textbf{Dropout}} \end{tabular}}
& \begin{tabular}[c]{@{}l@{}}{\fontsize{4.8}{5.2}\selectfont \textbf{C51. Static obstacle}} \\ {\fontsize{4.8}{5.2}\selectfont \textbf{navigation}}\end{tabular} & \begin{tabular}[c]{@{}l@{}}{\fontsize{4.5}{5}\selectfont ``Move forward and avoid collision''}\end{tabular} & \multirow{3}{*}{\includegraphics[height=1.4cm]{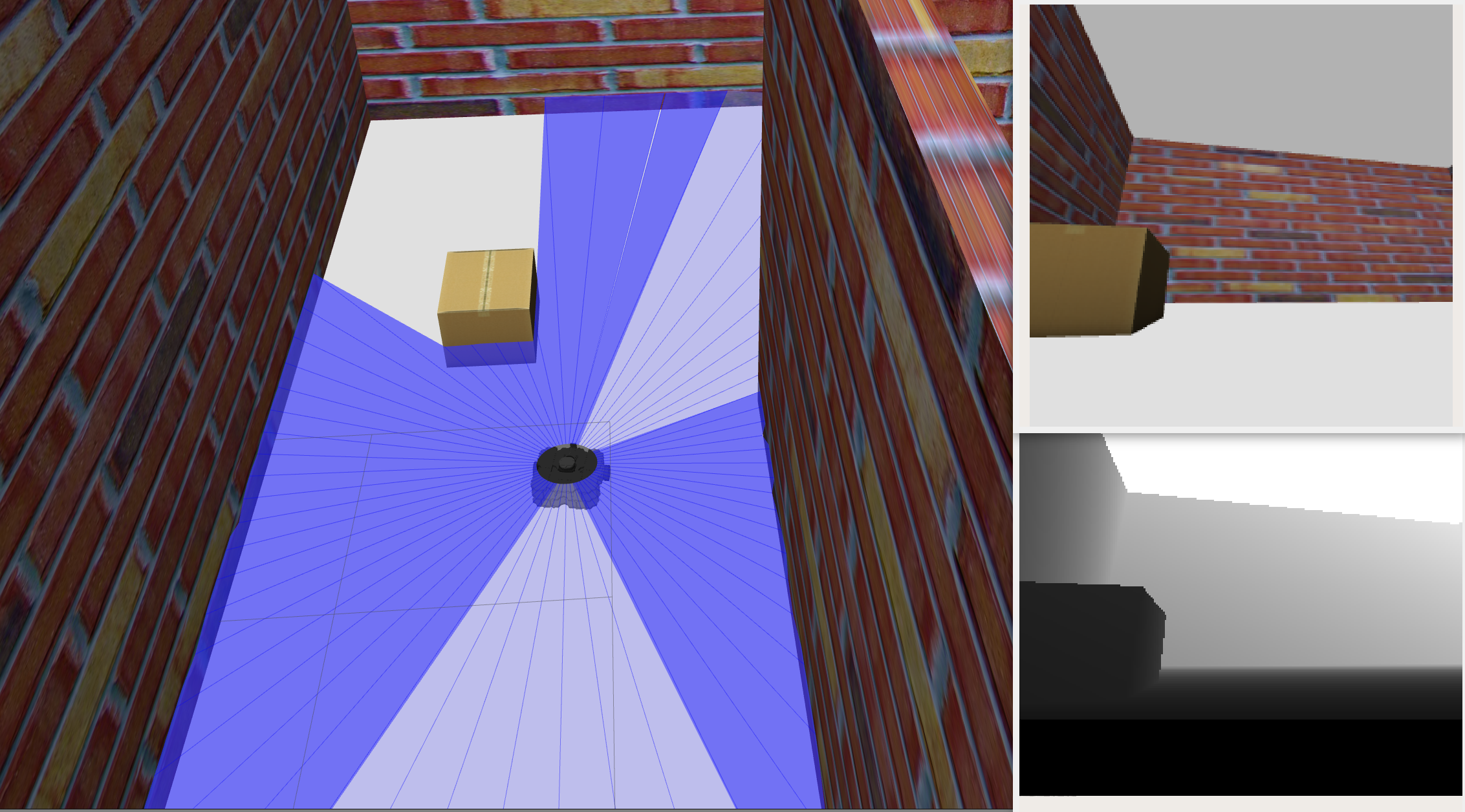}\hspace{1.5mm}\includegraphics[height=1.4cm]{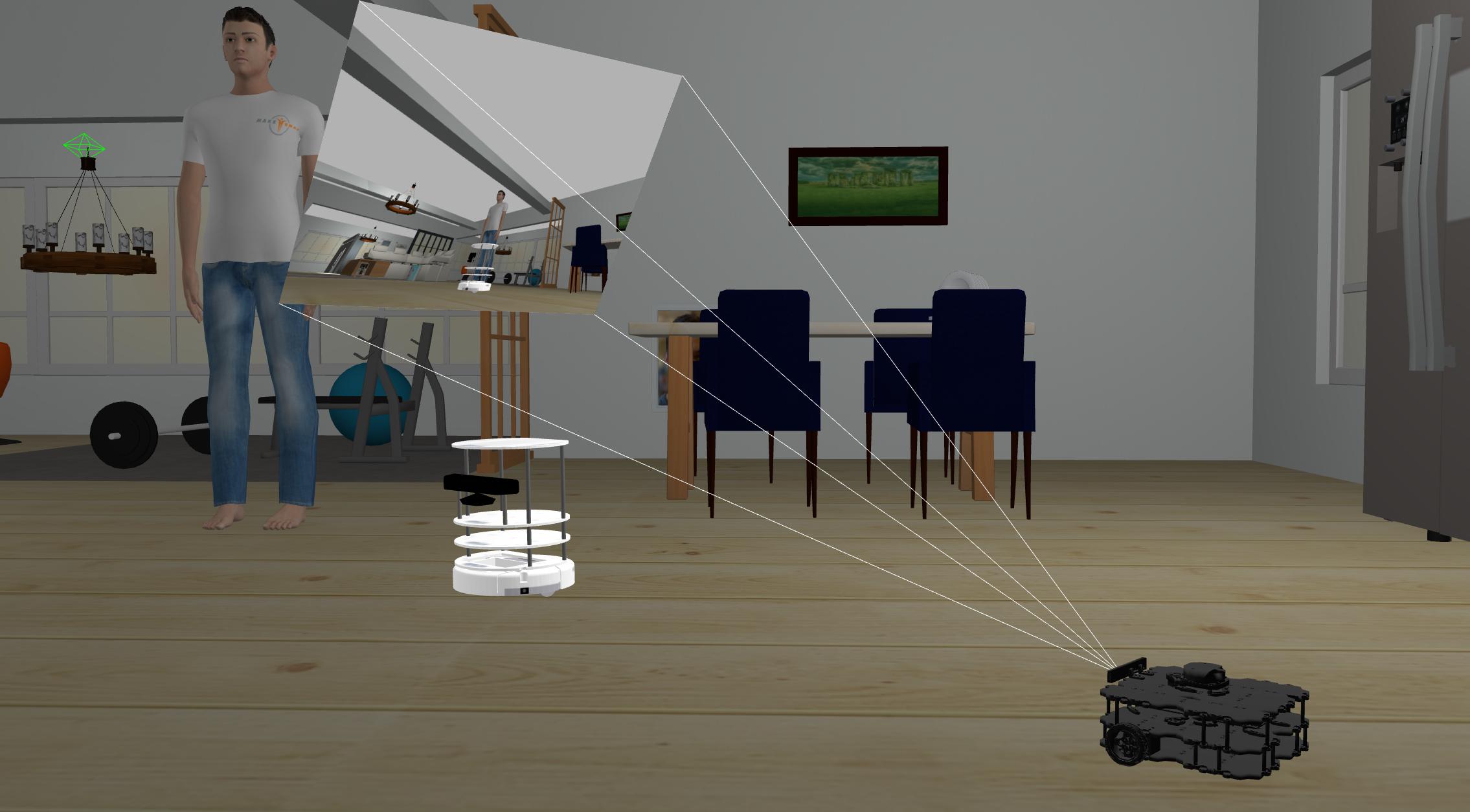}\hspace{1.5mm}\includegraphics[height=1.4cm]{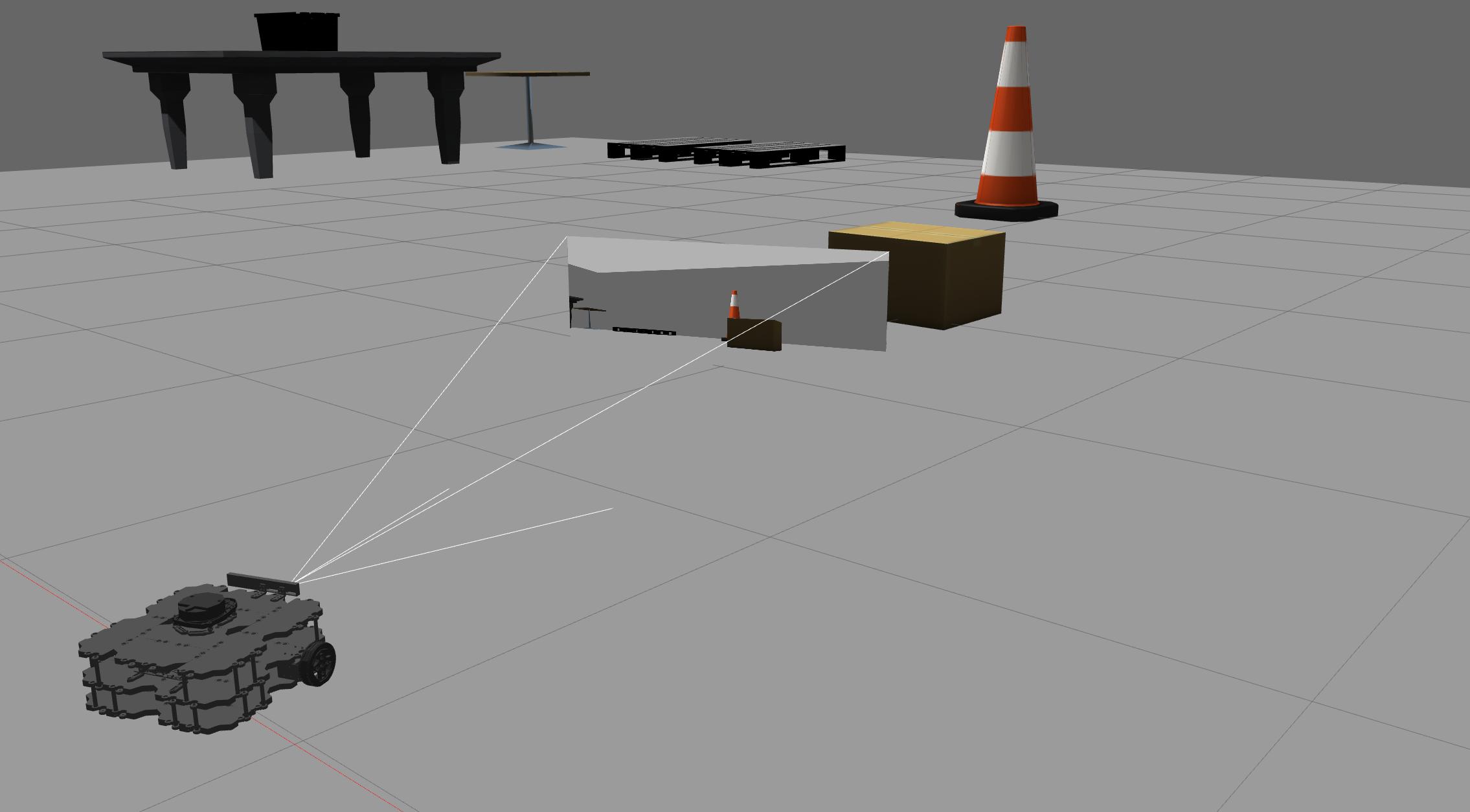}} \\
\cmidrule(lr){2-3}
& \begin{tabular}[c]{@{}l@{}}{\fontsize{4.8}{5.2}\selectfont \textbf{C52. Dynamic obstacle}} \\ {\fontsize{4.8}{5.2}\selectfont \textbf{navigation}}\end{tabular} & \begin{tabular}[c]{@{}l@{}}{\fontsize{4.5}{5}\selectfont ``Move forward and avoid} \\ {\fontsize{4.5}{5}\selectfont static and dynamic obstacles''}\end{tabular} & \\
\addlinespace[15pt]
\midrule

\end{tabular}%
}
\end{table*}

\subsection{Evaluated Models}
\label{sec:evaluated_models}
We evaluate five representative VLA models for mobile robot navigation: GR00T N1.6, \(\pi\)0.5, OpenVLA-OFT, ACT, and SmolVLA. All models are trained and evaluated with the same task definitions and success criteria, while model-specific fine-tuning choices, hardware, training time, and optimization settings are summarized in Table~\ref{tab:training_details}.

\subsection{Task Category Design}
\label{sec:tasks_category_design}
We evaluate generalization from ID to OOD conditions by varying task structure, environment, sensory inputs, and state/action distributions. These scenarios are designed to isolate different failure modes of VLA navigation policies, from low-level control and spatial extrapolation to semantic grounding, compositional behavior, and sensor robustness. The full set of benchmark tasks is summarized in Table~\ref{tab:tasks_categories}, while Table~\ref{tab:training_inputs_outputs} reports the sensor modalities, number of episodes, and episode lengths used for each task. We report these implementation details explicitly because, although they are often underreported in existing benchmark literature, such choices can materially affect policy training and evaluation. Documenting them provides practical guidance for reproducibility, fair comparison, and future implementations of the benchmark. For each task category, we define a separate set of OOD scenarios that perturb the aspects most relevant to that capability, and associate each scenario with a generalization domain such as visual, semantic, spatial, procedural, or modality-dropout generalization. We also identify which VLA axis is shifted relative to ID---vision, language, action/state, or a combination of these axes. The corresponding OOD settings and generalization domains are summarized in Table~\ref{tab:ood_settings}. For each category below, we first state the research question that motivated its inclusion in the benchmark. We then explain why its particular tasks were selected, which navigation capability they isolate, and how their OOD scenarios alter the training conditions to test that capability under distribution shift.

\begin{table}[!t]
\centering
\raggedright
\captionsetup{justification=raggedright,singlelinecheck=false,font=small}
\caption{Data collection details per task.}
\label{tab:training_inputs_outputs}
\newcommand{\tablefontsize}{\fontsize{4.5}{5.4}\selectfont}
\tablefontsize
\renewcommand{\arraystretch}{1.0}
\setlength{\tabcolsep}{0.5pt}
\resizebox{\linewidth}{!}{%
\begin{tabular}{@{}>{\tablefontsize}l>{\tablefontsize}l@{\hspace{1pt}}>{\tablefontsize}c@{\hspace{1pt}}>{\tablefontsize}c@{\hspace{1pt}}>{\tablefontsize}c@{\hspace{1pt}}>{\tablefontsize}c@{\hspace{1pt}}>{\tablefontsize}c@{\hspace{1pt}}>{\tablefontsize}c@{\hspace{1pt}}>{\tablefontsize}c@{\hspace{1pt}}>{\tablefontsize}c@{\hspace{1pt}}>{\tablefontsize}c@{}}
\toprule
{\tablefontsize\textbf{Tasks}} & {\tablefontsize\textbf{Details}} &
{\tablefontsize\textbf{C11}} & {\tablefontsize\textbf{C12}} & {\tablefontsize\textbf{C13}} &
{\tablefontsize\textbf{C21}} & {\tablefontsize\textbf{C31}} & {\tablefontsize\textbf{C41}} &
{\tablefontsize\textbf{C42}} & {\tablefontsize\textbf{C51}} & {\tablefontsize\textbf{C52}} \\
\midrule
\multirow{2}{*}{\tablefontsize\textbf{Proprioceptive State}}
& Robot's odometry (13 dim)
& \checkmark & \checkmark & \checkmark & \checkmark & \checkmark & \checkmark & \checkmark & \checkmark & \checkmark \\
& 2D LiDAR scan 
&  &  &  &  &  &  &  & \checkmark & \checkmark \\
\midrule
\multirow{3}{*}{\tablefontsize\textbf{Visual Input}}
& RGB ego camera
& \checkmark & \checkmark & \checkmark & \checkmark & \checkmark & \checkmark & \checkmark & \checkmark & \checkmark \\
& RGB world camera
&  &  &  &  &  & \checkmark &  &  &  \\[-1pt]
& Depth ego camera
&  &  &  &  &  &  &  & \checkmark & \checkmark \\
\midrule
\tablefontsize\textbf{Language Instruction} &  & \checkmark & \checkmark & \checkmark & \checkmark & \checkmark & \checkmark & \checkmark & \checkmark & \checkmark \\
\midrule
\tablefontsize\textbf{Action Space} & linear $x$, angular $z$
& \checkmark & \checkmark & \checkmark & \checkmark & \checkmark & \checkmark & \checkmark & \checkmark & \checkmark \\
\midrule
\tablefontsize\textbf{Number of episodes} &  & 100 & 100 & 160 & 2/100 & 10 & 1000 & 10 & 10 & 10 \\
\midrule
\tablefontsize\textbf{Episode length} & Control steps (20 Hz) & 400 & 400 & 500 & 400/12k & 500 & 500 & 12k & 12k & 12k \\
\bottomrule
\end{tabular}%
}

\end{table}

\textbf{Primitive Motion Control:  \textit{Q1. How well do models map metric language commands to precise actions?}}
The first category, primitive motion control, comprises basic discrete \textit{translational} and \textit{rotational} commands, including moving forward or backward by \(N\) meters and turning by \(N\) degrees, and \textit{compositions} of these motions.

\textit{OOD Scenarios:} The primary focus of this category is numeric generalization: we use unseen distances and rotation angles to test whether a model grounds numerical values in precise geometric motion rather than memorizing the magnitudes observed during training. These settings assess both interpolation between trained values and extrapolation beyond the trained numeric range. As complementary tests, new sequences of familiar translation and rotation commands evaluate whether individually learned motion primitives can be composed in an unseen order, while executing a familiar command in a new scene isolates whether visual changes disrupt the otherwise unchanged mapping from metric language and robot state to motion.

\textbf{Trajectory Extension: \textit{Q2. How far can models generalize beyond their training spatial horizon?}}
In this category, the robot follows a specified path, represented by a colored line, across two training settings: short-horizon paths, and longer-horizon paths. Training on different path/trajectory length exposes the model to different spatial span through the robot's geometric state space. Therefore, we trained the two sets of models on two different path lengths, one with a short spatial horizon and one with a longer spatial horizon.

\textit{OOD Senarios:} The primary challenge is maintaining reliable path-following behavior as the spatial extent increases beyond what was seen during training. So this setting tests whether models can interpolate to intermediate spatial states and extrapolate to extended, unseen ones. Since the robot operates from a local egocentric view, changes in the global path shape and length do not create a strong visual distribution shift in this task, and therefore the OOD scenarios mainly probe the models's ability to generalize in the state-action space, rather than the vision domain.

\begin{table}[h!]
\captionsetup{justification=raggedright,singlelinecheck=false,font=footnotesize}
\caption{OOD generalization shift categories (C1--C5).}
\label{tab:ood_settings}
\tiny
\renewcommand{\arraystretch}{0.85}
\setlength{\tabcolsep}{1pt}
\resizebox{\columnwidth}{!}{%
\begin{tabular}{@{}m{0.9cm}@{\hspace{4pt}}p{1.35cm}@{\hspace{4pt}}p{1.45cm}@{\hspace{4pt}}p{3.2cm}@{}}
\toprule
\textbf{Category} & \textbf{Axes} & \textbf{OOD Settings} & \textbf{Example Command} \\
\midrule

\multirow{3}{*}[-2pt]{\textbf{C1}}
& \cellcolor{green!10}{\fontsize{4.8}{5.2}\selectfont \textbf{A-L}} & \cellcolor{green!10}{\fontsize{4.8}{5.2}\selectfont \textbf{OOD Numeric}} & \cellcolor{green!10}{\fontsize{4.5}{5}\selectfont ``Move 2.5 meters forward''} \\
\addlinespace[2pt]
& \cellcolor{red!10}{\fontsize{4.8}{5.2}\selectfont \textbf{L-A}} & \cellcolor{red!10}{\fontsize{4.8}{5.2}\selectfont \textbf{OOD Sequence}} & \cellcolor{red!10}{\fontsize{4.5}{5}\selectfont ``Move 2 meters and then rotate 10 degrees''} \\
\addlinespace[2pt]
& \cellcolor{yellow!25}{\fontsize{4.8}{5.2}\selectfont \textbf{V}} & \cellcolor{yellow!25}{\fontsize{4.8}{5.2}\selectfont \textbf{OOD Scene}} & \cellcolor{yellow!25}{\fontsize{4.5}{5}\selectfont ``Move 2 meters forward''} \\
\addlinespace[2pt]
\bottomrule
\multirow{2}{*}[-3pt]{\textbf{C2}}
& \cellcolor{green!10}{\fontsize{4.8}{5.2}\selectfont \textbf{A-V}} & \cellcolor{green!10}{\fontsize{4.8}{5.2}\selectfont \textbf{OOD Length(short)}} & \cellcolor{green!10}\begin{tabular}[c]{@{}l@{}}{\fontsize{4.5}{5}\selectfont ``Follow the red line''}\end{tabular} \\
\addlinespace[2pt]
& \cellcolor{green!10}{\fontsize{4.8}{5.2}\selectfont \textbf{A-V}} & \cellcolor{green!10}{\fontsize{4.8}{5.2}\selectfont \textbf{OOD Length(long)}} & \cellcolor{green!10}\begin{tabular}[c]{@{}l@{}}{\fontsize{4.5}{5}\selectfont ``Follow the red line''}\end{tabular} \\
\addlinespace[2pt]

\bottomrule
\multirow{2}{*}[-2pt]{\textbf{C3}}
& \cellcolor{red!10}{\fontsize{4.8}{5.2}\selectfont \textbf{L-A}} & \cellcolor{red!10}{\fontsize{4.8}{5.2}\selectfont \textbf{OOD L-Sequence}} & \cellcolor{red!10}\begin{tabular}[c]{@{}l@{}}{\fontsize{4.5}{5}\selectfont ``Navigate to the chair, then cone,}{\fontsize{4.5}{5}\selectfont and then person''}
\end{tabular} \\
\addlinespace[2pt]
& \cellcolor{yellow!25}{\fontsize{4.8}{5.2}\selectfont \textbf{V}} & \cellcolor{yellow!25}{\fontsize{4.8}{5.2}\selectfont \textbf{Visual memory}} & \cellcolor{yellow!25}\begin{tabular}[c]{@{}l@{}}{\fontsize{4.5}{5}\selectfont ``Navigate to the cone, then chair,} {\fontsize{4.5}{5}\selectfont and then person''}
\end{tabular} \\
\bottomrule

\multirow{7}{*}[-6pt]{\textbf{C4}}
& \cellcolor{yellow!25}{\fontsize{4.8}{5.2}\selectfont \textbf{V}} & \cellcolor{yellow!25}{\fontsize{4.8}{5.2}\selectfont \textbf{OOD Init Pose}} & \cellcolor{yellow!25}{\fontsize{4.5}{5}\selectfont ``Navigate to the cone''} \\
\addlinespace[2pt]
& \cellcolor{yellow!25}{\fontsize{4.8}{5.2}\selectfont \textbf{V-A}} & \cellcolor{yellow!25}{\fontsize{4.8}{5.2}\selectfont \textbf{OOD Target Pose}} & \cellcolor{yellow!25}{\fontsize{4.5}{5}\selectfont ``Navigate to the cone''} \\
\addlinespace[2pt]
& \cellcolor{yellow!25}{\fontsize{4.8}{5.2}\selectfont \textbf{V}} &  \cellcolor{yellow!25}{\fontsize{4.8}{5.2}\selectfont \textbf{OOD Scene}} & \cellcolor{yellow!25}{\fontsize{4.5}{5}\selectfont ``Navigate to the cone''} \\
\addlinespace[2pt]

& \cellcolor{yellow!25}{\fontsize{4.8}{5.2}\selectfont \textbf{V}} &  \cellcolor{yellow!25}{\fontsize{4.8}{5.2}\selectfont \textbf{OOD Cam View}} & \cellcolor{yellow!25}{\fontsize{4.5}{5}\selectfont ``Navigate to the cone''} \\
\addlinespace[2pt]
& \cellcolor{blue!10}{\fontsize{4.8}{5.2}\selectfont \textbf{V-L}} & \cellcolor{blue!10}{\fontsize{4.8}{5.2}\selectfont \textbf{OOD Object}} & \cellcolor{blue!10}{\fontsize{4.5}{5}\selectfont ``Navigate to the sofa''/"Follow the wheelchair"} \\
\addlinespace[2pt]
& \cellcolor{blue!10}{\fontsize{4.8}{5.2}\selectfont \textbf{V-L}} & \cellcolor{blue!10}{\fontsize{4.8}{5.2}\selectfont \textbf{OOD L-Instruction}} & \cellcolor{blue!10}{\fontsize{4.5}{5}\selectfont ``Navigate to person''/"Navigate to human"} \\
\addlinespace[2pt]
& \cellcolor{blue!10}{\fontsize{4.8}{5.2}\selectfont \textbf{V-L}} & \cellcolor{blue!10}\begin{tabular}[c]{@{}l@{}}{\fontsize{4.8}{5.2}\selectfont \textbf{World Knowledge}} \\ \end{tabular} & \cellcolor{blue!10}{\fontsize{4.5}{5}\selectfont ``Navigate to the Finland's flag''} \\
\midrule

\multirow{3}{*}[-2pt]{\textbf{C5}}
& \cellcolor{gray!20}{\fontsize{4.8}{5.2}\selectfont \textbf{V}} & \cellcolor{gray!20}\begin{tabular}[c]{@{}l@{}}{\fontsize{4.8}{5.2}\selectfont \textbf{RGB Dropout}} \\ \end{tabular} & \cellcolor{gray!20}{\fontsize{4.5}{5}\selectfont ``Move forward and avoid collision''} \\
\addlinespace[2pt]
& \cellcolor{gray!20}{\fontsize{4.8}{5.2}\selectfont \textbf{V}} & \cellcolor{gray!20}\begin{tabular}[c]{@{}l@{}}{\fontsize{4.8}{5.2}\selectfont \textbf{Depth Dropout}} \\ \end{tabular} & \cellcolor{gray!20}{\fontsize{4.5}{5}\selectfont ``Move forward and avoid collision''} \\
\addlinespace[2pt]
& \cellcolor{gray!20}{\fontsize{4.8}{5.2}\selectfont \textbf{A}} & \cellcolor{gray!20}\begin{tabular}[c]{@{}l@{}}{\fontsize{4.8}{5.2}\selectfont \textbf{Lidar Dropout}} \\ \end{tabular} & \cellcolor{gray!20}{\fontsize{4.5}{5}\selectfont ``Move forward and avoid collision''} \\

\addlinespace[2pt]
\midrule
\multicolumn{4}{l}{{\fontsize{4.8}{5.2}\selectfont \textbf{Generalization Domain:}}\hspace{6pt}%
  \colorbox{green!10}{\fontsize{4.5}{5}\selectfont\strut~\textbf{Spatial}~}~
  \colorbox{red!10}{\fontsize{4.5}{5}\selectfont\strut~\textbf{Procedural}~}~
  \colorbox{yellow!25}{\fontsize{4.5}{5}\selectfont\strut~\textbf{Visual}~}~
  \colorbox{blue!10}{\fontsize{4.5}{5}\selectfont\strut~\textbf{Semantic}~}~
  \colorbox{gray!20}{\fontsize{4.5}{5}\selectfont\strut~\textbf{Modality Dropout}~}%
} \\
\bottomrule

\end{tabular}%
}%

\end{table}

\begin{figure*}[!t]
\centering
\includegraphics[
  width=\textwidth,
  height=0.22\textheight
]{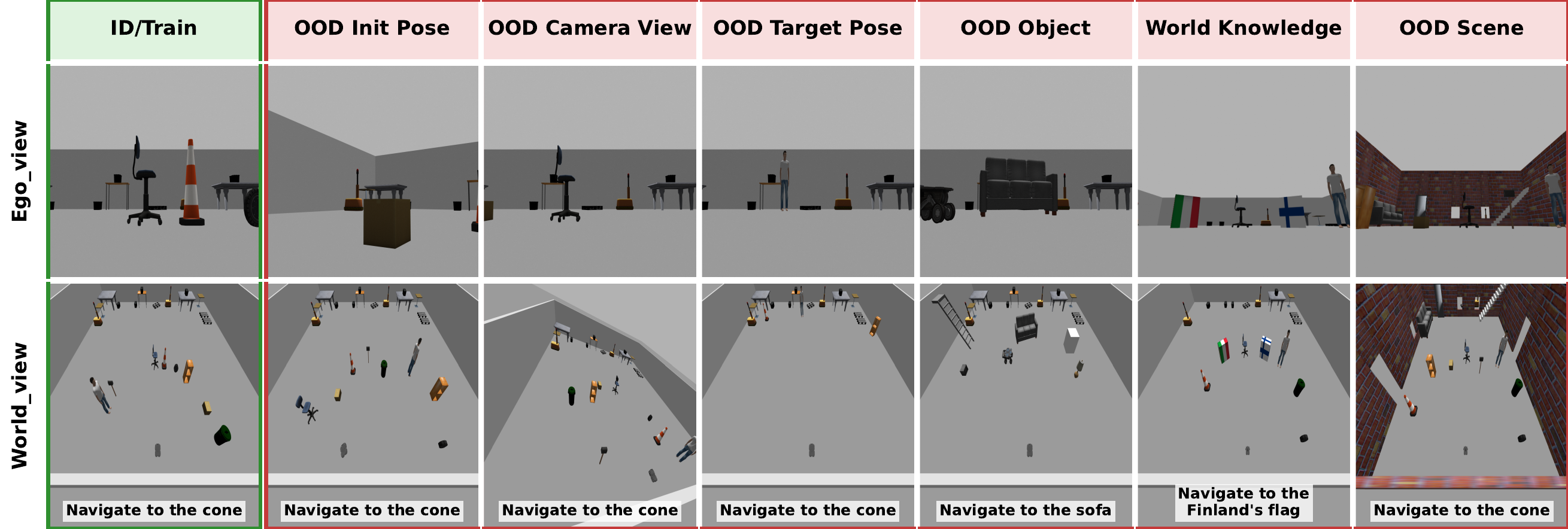}
\caption{ID and OOD ego/world-view examples for semantic goal grounding.}
\label{fig:c4_id_ood}
\end{figure*}

\textbf{Compositional Tasks:  \textit{Q3. Can models interpret new sequence ordered instructions?
 Q4. Do they maintain a visual memory over the task?}}
In this category, the robot must visit three distinct objects in a specified order. The model must ground the language sequence in the initial visual layout and retain a short-term visual memory of where each object is located relative to the others and to the robot, even after they disappear in the egocentric camera view.

\textit{OOD Scenarios:} We test two sequence shifts using the same familiar objects. OOD language sequence changes the commanded visit order, probing whether the model can generalize the procedural order in the instruction or instead treats each sequence as a fixed memorized task, similar to new command compositions in primitive motion control task. Also, OOD visual sequence changes the initial spatial order of the objects, testing whether the model can bind instruction order to visual memory.

\textbf{Semantic Goal Grounding: \textit{Q5. Can VLAs turn visual-language recognition into embodied navigation?}}
This category is closest to the object-centric evaluations commonly used in manipulation benchmarks, where models must connect language to a visual target. Here, we test the same general capability in mobile navigation, where the target must be maintained as an embodied navigation objective throughout closed-loop execution. We consider two main tasks: goal-directed navigation, where the robot navigates to one of 10 training objects using both egocentric and world-view images, and dynamic target following, where the robot follows a moving target such as a person, nurse, or man in a red shirt.

\textit{OOD Scenarios:} In this category, we focus on varying the robot initial pose, target pose, scene, camera viewpoint, and target identity; representative examples of these ID and OOD conditions are shown in Fig.~\ref{fig:c4_id_ood}. Motivated by external top-down views used in manipulation setups and in shared workspaces such as warehouses, we include a world-camera modality to evaluate how navigation policies use global visual information. We then substantially change its viewpoint to test whether the policy remains robust or relies on a fixed global observation. The world-knowledge setting further tests whether the model can recognize and use internet-scale visual-language knowledge inherited from VLM pretraining. 

\textbf{Safety Constraint Modality Dropout: \textit{Q6. How do added sensors affect safe navigation performance, and how fragile are VLAs under sensor dropout?}}
The primary inputs to VLA policies are typically RGB images, language instructions, and the robot's proprioceptive state. In this category, we augment these inputs with egocentric depth images and 2D LiDAR in the proprioceptive state, as geometric sensing plays an essential role in obstacle avoidance, a core mobile-navigation problem. Using static and dynamic obstacle-navigation tasks, we test whether these additional modalities improve safety and reduce collisions. The richer modality set also enables controlled analysis of each input's contribution to safe behavior and reveals whether a policy benefits from, ignores, or becomes overly dependent on a particular sensor.

\textit{OOD Scenarios:} We use modality dropout as a controlled disruption to the trained input space by applying a permissive mask to one selected modality at a time. This masks RGB, depth, or LiDAR evidence from the corresponding input stream without changing the task instruction or controller output space. We evaluate these settings to measure each model's reliance on different sensors and whether it can still maintain safe behavior when one, or more, modalities are masked.

\subsection{Generalization Domains}
Although each OOD scenario defined above must be evaluated separately within its corresponding task, several scenarios introduce related shifts and probe the same underlying generalization capability across different tasks. We therefore also group these task-specific cases into five cross-task generalization domains--visual, semantic, spatial, procedural, and modality dropout--which enable joint analysis of scenarios that stress a common aspect of the model; Table~\ref{tab:ood_settings} shows how the individual OOD scenarios are organized into these five domains. Together, these domains systematically probe the three core components of a VLA--vision, language, and action/state--beyond the ID settings and reveal which components are most sensitive to distribution shift. Below, we describe each domain in detail.

\begin{figure*}[!t]
\centering
\includegraphics[width=0.8\textwidth, height=0.2\textheight]{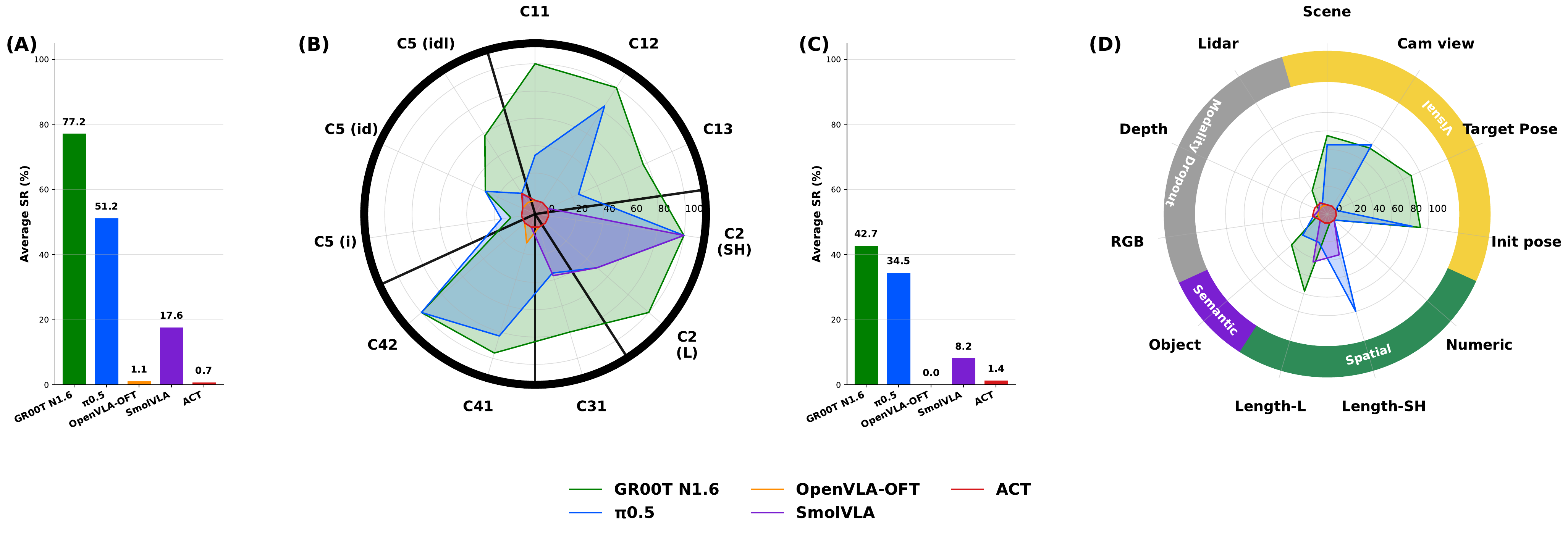}
\caption{Success rate for ID benchmark tasks and OOD generalization domains.}
\label{fig:id-success-radar}
\end{figure*}

\textbf{Visual Generalization}
Visual generalization targets shifts in the image distribution, such as changes in scene, camera viewpoint, robot initial pose or the initial visual order of objects. These settings are designed to primarily stress the visual backbone of the VLA while keeping the language instruction and state/action space fixed.

\textbf{Semantic Generalization}
Semantic generalization targets joint vision-language grounding. By changing target objects, language instructions, or using world-knowledge goals, it tests whether the model can map language to the correct visual concept rather than rely on memorized target.

\textbf{Spatial Generalization}
Spatial generalization targets action-generation component's understanding of space. The main shift is unseen trajectory lengths, testing whether the control policy extends beyond the spatial states or horizon seen in demonstrations.

\textbf{Procedural Generalization}
Procedural generalization evaluates whether models can compose familiar tasks into new ordered behaviors and tasks. It tests whether the model grounds the procedure across language, vision, and action, or treats each demonstrated order as a memorized task.

\textbf{Modality Dropout Generalization}
Modality dropout generalization first studies how VLAs incorporate navigation-specific modalities, such as depth and LiDAR, beyond the standard visual and robot-state inputs. It then removes RGB, depth, or LiDAR inputs to disrupt different VLA components and test whether safe obstacle avoidance remains possible without each modality.

\section{Results}
\label{sec:results}
\begin{figure*}[!t]
\begin{minipage}[t]{0.495\textwidth}
\vspace{0pt}
\centering
\subfloat[A\label{fig:dynamic-hospital}]{%
\includegraphics[width=0.49\linewidth]{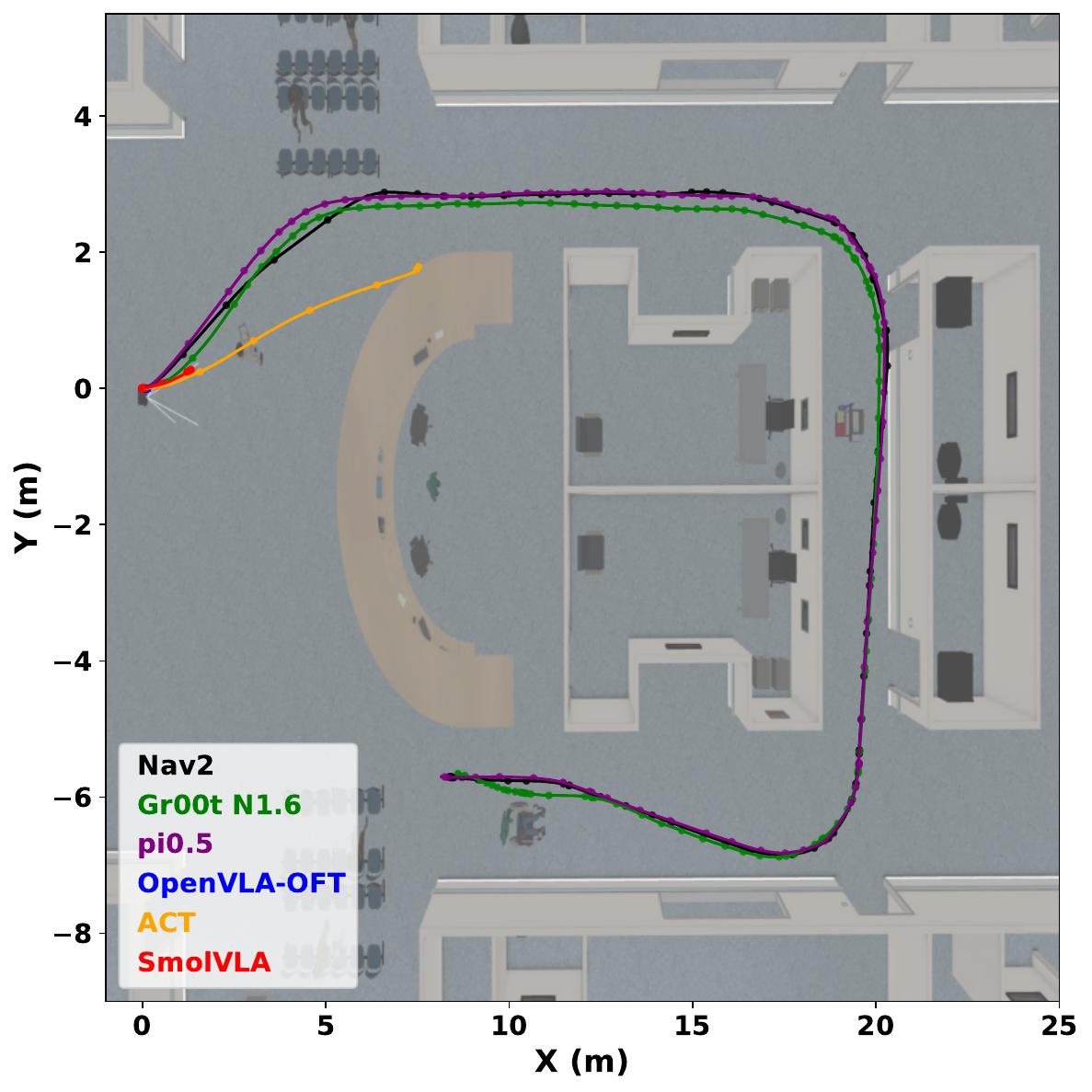}}
\hfill
\subfloat[B\label{fig:dynamic-bookstore}]{%
\includegraphics[width=0.49\linewidth]{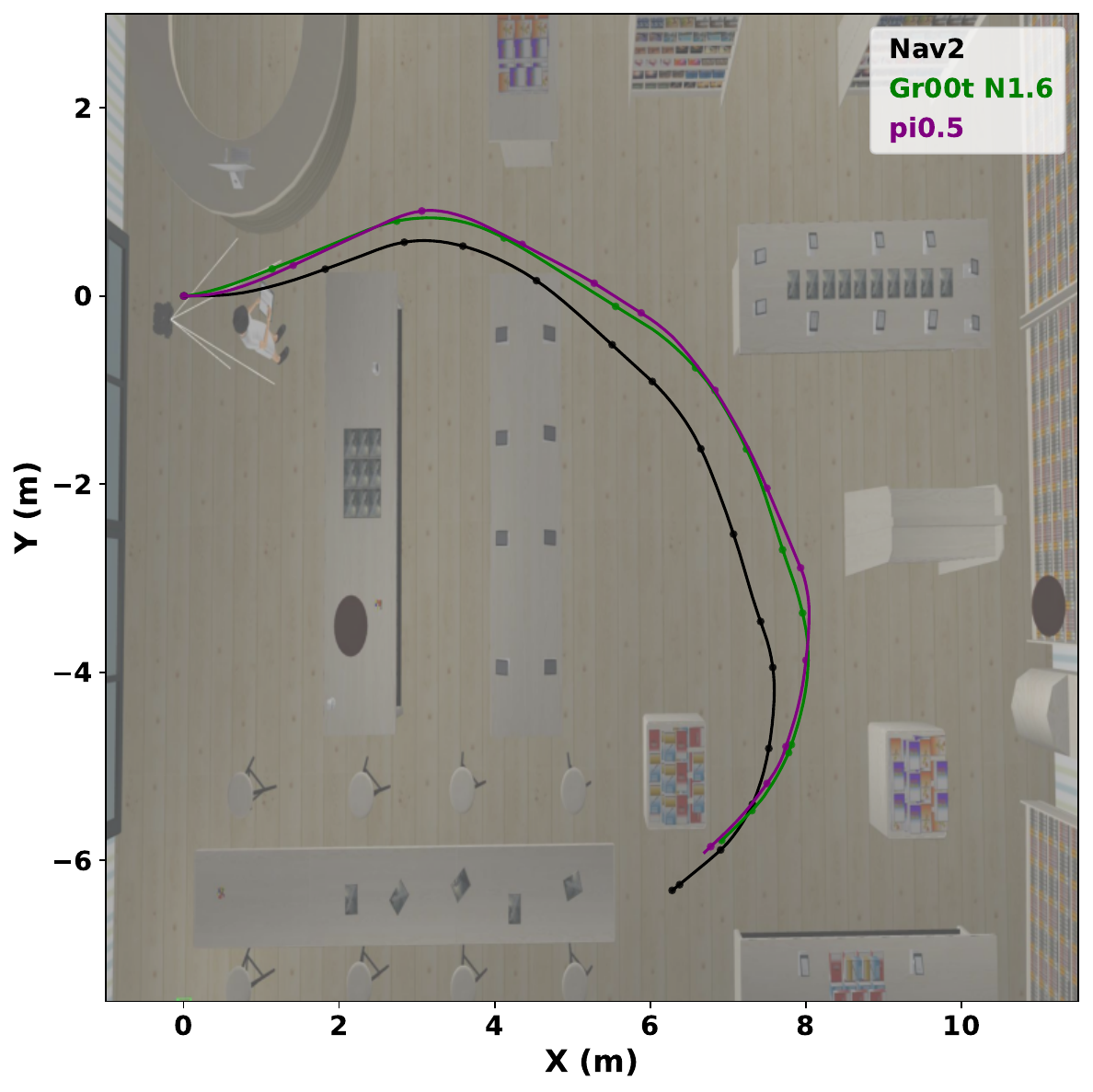}}
\caption{Trajectory examples for dynamic target following in the ID hospital scene and OOD bookstore scene.}
\label{fig:dynamic-interaction}
\end{minipage}%
\hfill%
\begin{minipage}[t]{0.495\textwidth}
\vspace{0pt}
\centering
\includegraphics[width=\linewidth]{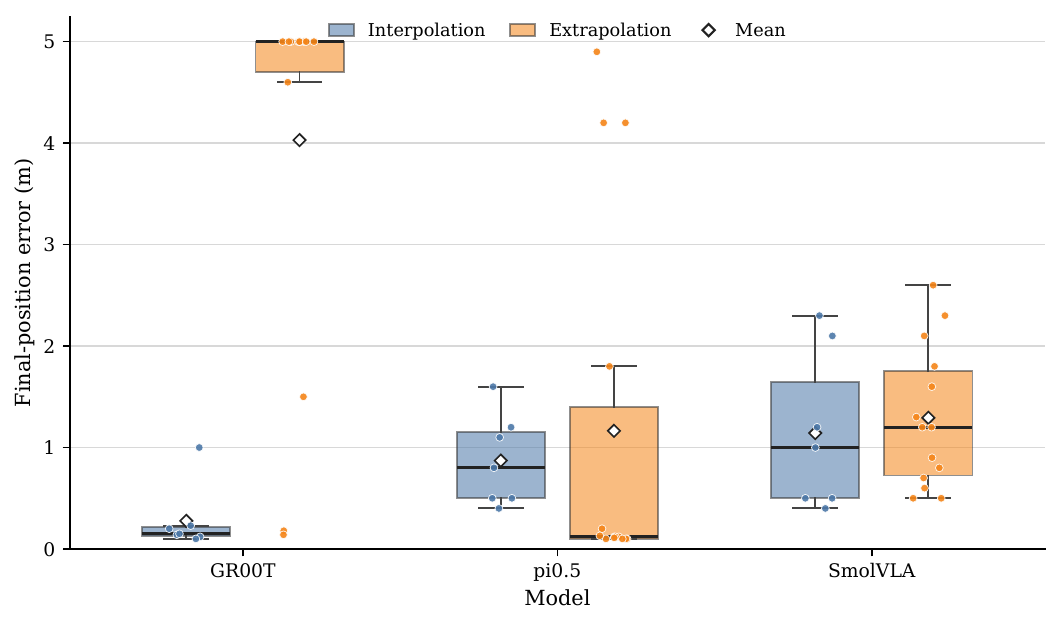}
\caption{Trajectory-extension performance under short- and long-horizon training.}
\label{fig:interpolation_extrapolation_boxplot}
\end{minipage}
\end{figure*}

This section presents the main quantitative and qualitative results obtained on the unified benchmark.

\textbf{ID Performance.}
Although this work focuses on generalization, ID performance is also critical: a model must first solve the training-distribution version of each task before OOD robustness is meaningful. We therefore report task-wise ID success rate (SR) for each model in Fig.~\ref{fig:id-success-radar}(A, B), using the SR definition summarized in Table~\ref{tab:id-success-task-metrics}.
\begin{table}[h!]
\centering
\caption{SR metric definitions across benchmark tasks.}
\label{tab:id-success-task-metrics}
\scriptsize
\renewcommand{\arraystretch}{1.15}
\setlength{\tabcolsep}{3pt}
\begin{tabularx}{\linewidth}{@{}lX@{}}
\toprule
\textbf{Task} & \textbf{SR metric} \\
\midrule
\textbf{C11} & Final position error is less than 0.5 m from the expected position. \\
\textbf{C12} & Final heading error is less than 5 degrees from the expected orientation. \\
\textbf{C13} & Final position error is less than 0.5 m and final heading error is less than 5 degrees. \\
\textbf{C21} & The robot stops automatically at the end of the line with less than 1 m final position error. \\
\textbf{C31} & The robot visits the target objects in the instructed order and stops automatically at the last target. \\
\textbf{C41} & The robot navigates to the correct object and stops automatically without colliding with the target object. \\
\textbf{C42} & The robot follows the dynamic target until the end of the run without colliding with it. \\
\textbf{C51} & The robot keeps moving until the end of the run without collision. \\
\textbf{C52} & The robot keeps moving until the end of the run without collision. \\
\bottomrule
\end{tabularx}
\end{table}

Overall, GR00T N1.6 achieves the strongest ID performance across nearly all tasks, followed by \(\pi\)0.5; SmolVLA succeeds only on a subset, while ACT and OpenVLA-OFT remain weak. Notably, GR00T maintains high ID performance even on tasks trained with limited demonstration data, indicating comparatively strong data efficiency. C5 remains the most difficult ID setting because success requires both forward progress and collision avoidance. The effect of additional geometric sensing is strongly model-dependent. For GR00T, SR increases from 8\% with RGB alone to 30\% with RGB--depth and 58\% with RGB--depth--LiDAR, indicating cumulative gains and the largest improvement from adding LiDAR. For \(\pi\)0.5, depth raises SR from 15\% to 30\%, whereas adding LiDAR reduces it to 8\%, showing that providing more modalities does not necessarily result in better sensor integration. ACT improves only when LiDAR is available, from 0\% to 8\%, while SmolVLA changes only marginally, reaching 7\% with the full modality set; OpenVLA-OFT remains at 0\% in all three settings.

\textbf{Quantitative OOD Generalization Summary.}
Fig.~\ref{fig:id-success-radar}(C, D) reports OOD success rates averaged over the common distribution shifts grouped under each generalization domain.

\textit{Visual.} Visual shifts are the most tolerable domain for GR00T, which remains above roughly 70\% SR across scene, camera-view, initial-pose, and target-pose changes; Fig.~\ref{fig:dynamic-interaction} shows representative performance in a new-scene environment. Overall, \(\pi\)0.5 achieves the second-best performance in this domain, while the remaining models, which already fail to establish reliable performance under ID conditions, also fail as expected under OOD visual shifts.

\textit{Spatial.} None of the models generalize well to unseen numeric commands, indicating weak grounding between metric language and precise state-action execution. For trajectory extension, Fig.~\ref{fig:interpolation_extrapolation_boxplot} shows that GR00T depends strongly on the spatial-horizon diversity of the demonstrations: it achieves the highest interpolation accuracy, consistent with its strong ID performance, but its error increases substantially when extrapolating to longer horizons. \(\pi\)0.5 has a low extrapolation error, indicating that it extends beyond the demonstrated spatial horizon more effectively than the other models in a substantial portion of the trials. SmolVLA exhibits a smaller gap between interpolation and extrapolation, indicating more consistent performance across the two regimes.

\textit{Semantic.} Semantic shifts, especially unseen target objects, are difficult for all models. This indicates that current generalist VLAs remain narrow in the object identities they can reliably ground for navigation, despite the diversity already present in the ID setting.

\begin{figure*}[!t]
\centering
\includegraphics[width=\textwidth]{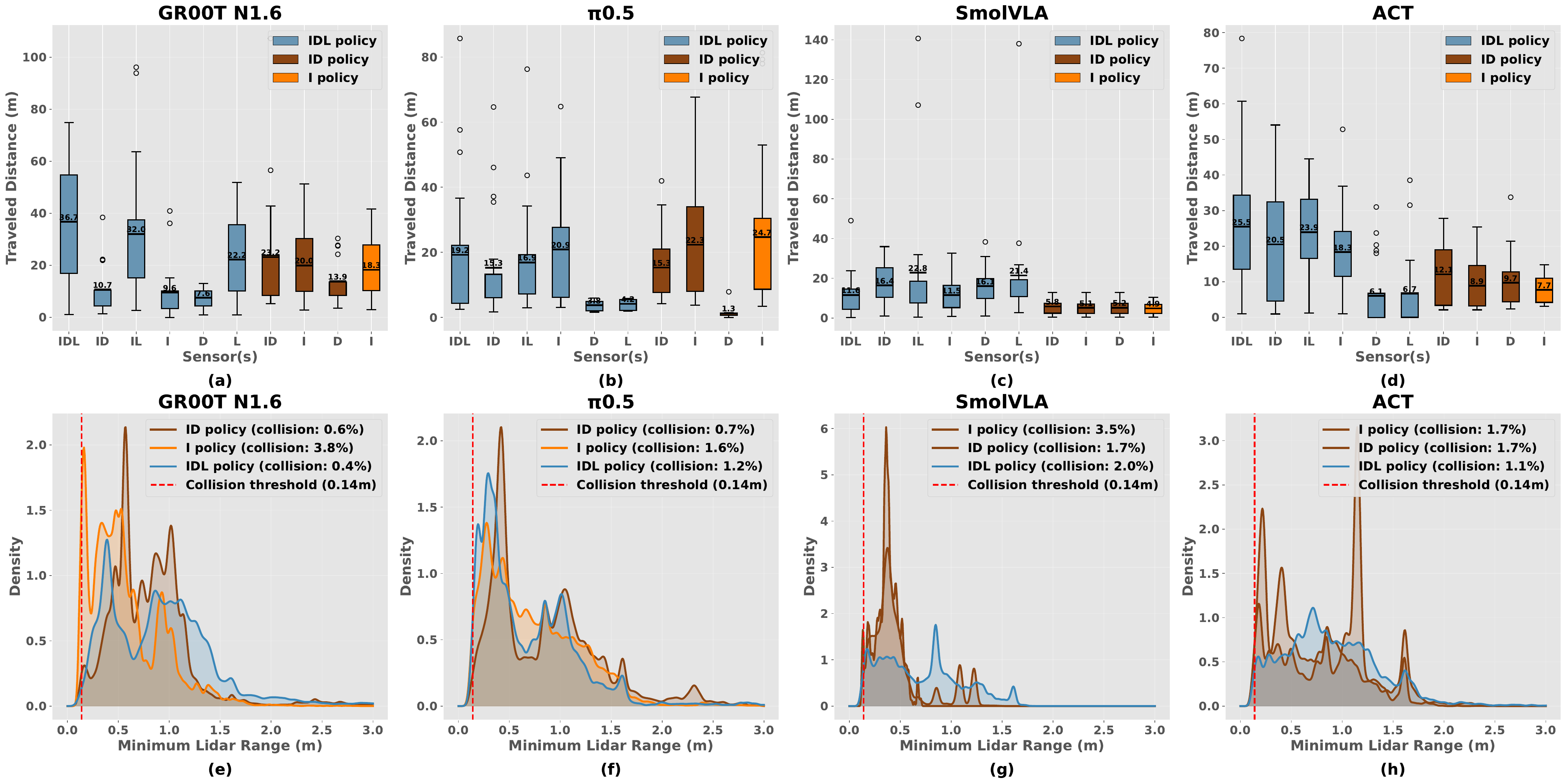}
\caption{Comparison across input-modality settings.}
\label{fig:modalities}
\end{figure*}

\textit{Modality Dropout.} Fig.~\ref{fig:modalities} compares RGB-only (I), RGB-depth (ID), and RGB-depth-LiDAR (IDL) policies under modality dropout, measuring both how far the robot can travel and how often it collides. The first row compares travelled distance under equal trials and shows whether each policy can continue moving safely when a modality is masked. The second row reports collision rate when all modalities for a policy are available, showing whether adding depth or LiDAR improves safety before dropout is applied.

For GR00T, adding LiDAR gives the clearest gain: the IDL policy travels farther and collides less than the image-only and image-depth policies. When LiDAR is masked, performance drops sharply, often through collision or early stopping, which indicates strong reliance on the proprioceptive/geometric state channel. Depth alone has a smaller effect for GR00T, suggesting that its obstacle-avoidance behavior is driven more by LiDAR than by visual geometry.

For \(\pi\)0.5, LiDAR does not provide the same improvement. The RGB-only policy remains competitive in travelled distance, indicating stronger reliance on visual input, while depth improves safety by reducing collisions even if it can reduce progress. This suggests that \(\pi\)0.5 uses depth more as a conservative safety cue than as a driver of faster navigation.

For SmolVLA, the policies sometimes move partway through the route but do not show a clear single-modality dependence comparable to GR00T or \(\pi\)0.5. Their behavior suggests weaker integration of the added navigation sensors rather than a consistent reliance on RGB, depth, or LiDAR. For ACT, adding LiDAR also reduces collision rate and increases travelled distance, indicating that ACT benefits from the extra geometric sensing even though its overall performance remains limited.

\begin{table}[!t]
\centering
\captionsetup{justification=raggedright,singlelinecheck=false,font=scriptsize}
\caption{Qualitative OOD capability summary.}
\label{tab:ood-capability-summary}
\scriptsize
\renewcommand{\arraystretch}{1.05}
\setlength{\tabcolsep}{1pt}
\begin{tabularx}{\columnwidth}{@{}>{\raggedright\arraybackslash}p{2.0cm}@{\hspace{0.5pt}}*{5}{>{\centering\arraybackslash}X}@{}}
\toprule
\fontsize{6.2}{6.5}\selectfont\textbf{Capability} & \fontsize{6.2}{6.5}\selectfont\textbf{GR00T N1.6} & \fontsize{6.2}{6.5}\selectfont\textbf{\boldmath$\pi$0.5} & \fontsize{6.2}{6.5}\selectfont\textbf{OpenVLA-OFT} & \fontsize{6.2}{6.5}\selectfont\textbf{ACT} & \fontsize{6.2}{6.5}\selectfont\textbf{SmolVLA} \\
\midrule
\cellcolor{red!10}\fontsize{6.2}{6.5}\selectfont\textbf{OOD Sequence} & \cellcolor{red!10}\xmark & \cellcolor{red!10}\cmark & \cellcolor{red!10}\xmark & \cellcolor{red!10}\xmark & \cellcolor{red!10}\xmark \\
\cellcolor{yellow!25}\fontsize{6.2}{6.5}\selectfont\textbf{Visual Memory} & \cellcolor{yellow!25}\xmark & \cellcolor{yellow!25}\xmark & \cellcolor{yellow!25}\xmark & \cellcolor{yellow!25}\xmark & \cellcolor{yellow!25}\xmark \\
\cellcolor{blue!10}\fontsize{6.2}{6.5}\selectfont\textbf{World Knowledge} & \cellcolor{blue!10}\xmark & \cellcolor{blue!10}\xmark & \cellcolor{blue!10}\xmark & \cellcolor{blue!10}\xmark & \cellcolor{blue!10}\xmark \\
\cellcolor{blue!10}\fontsize{6.2}{6.5}\selectfont\textbf{OOD L-Inst. Simple} & \cellcolor{blue!10}\cmark & \cellcolor{blue!10}\cmark & \cellcolor{blue!10}\xmark & \cellcolor{blue!10}\xmark & \cellcolor{blue!10}\xmark \\
\cellcolor{blue!10}\fontsize{6.2}{6.5}\selectfont\textbf{OOD L-Inst. Complex} & \cellcolor{blue!10}\cmark & \cellcolor{blue!10}\xmark & \cellcolor{blue!10}\xmark & \cellcolor{blue!10}\xmark & \cellcolor{blue!10}\xmark \\
\bottomrule
\end{tabularx}
\end{table}
\textbf{Qualitative OOD Capability Summary}
Table~\ref{tab:ood-capability-summary} summarizes qualitative OOD capabilities. In contrast to its weaker ID performance on compositional tasks, \(\pi\)0.5 shows the promising procedural generalization: it follows new target orders, even when it does not complete the full sequence. However, none of the evaluated models demonstrates reliable visual memory. Once a previously observed object leaves the egocentric view, the policies show no clear evidence that they retain or use its earlier location, even though the language instruction identifies the relevant objects and specifies their visitation order. Their subsequent behavior therefore appears to depend primarily on currently visible evidence rather than information preserved from earlier frames. All models fail the world-knowledge setting. For language shifts, GR00T and \(\pi\)0.5 handle simple paraphrases such as ``person''/``human'', while only GR00T handles more complex descriptions such as referring to a chair as "the object to sit on".

Table~\ref{tab:question-model-summary} summarizes the main finding for each model under the six research questions that motivate the benchmark tasks.
\begin{table*}[!t]
\centering
\captionsetup{justification=raggedright,singlelinecheck=false,font=scriptsize}
\caption{Question-wise summary of the key findings for each evaluated model.}
\label{tab:question-model-summary}
\scriptsize
\renewcommand{\arraystretch}{1.15}
\setlength{\tabcolsep}{2.5pt}
\begin{tabularx}{\textwidth}{@{}>{\raggedright\arraybackslash}p{2.35cm}*{5}{>{\raggedright\arraybackslash}X}@{}}
\toprule
\textbf{Research question} & \textbf{GR00T N1.6} & \textbf{\boldmath$\pi$0.5} & \textbf{OpenVLA-OFT} & \textbf{SmolVLA} & \textbf{ACT} \\
\midrule
\textbf{Q1: Numeric motion grounding} & Strongest ID control; fails unseen numeric values & Mixed ID accuracy; fails unseen numeric values & Fails ID and OOD commands & Fails most ID and OOD commands & Fails ID and OOD commands \\
\textbf{Q2: Spatial-horizon extension} & Best interpolation; degrades under extrapolation & Low extrapolation error; strongest extension & Largely fails & Partial but consistent generalization & Largely fails \\
\textbf{Q3: Procedural order} & Strong ID; fails new target orders & Best generalization to new orders, but often incomplete & Fails & Limited ID; fails new orders & Fails \\
\textbf{Q4: Visual memory} & No clear visual memory & No clear visual memory & No clear visual memory & No clear visual memory & No clear visual memory \\
\textbf{Q5: Semantic navigation} & Strong ID and visual robustness; fails unseen semantics & Second-best ID/visual performance; weak unseen semantics & Weak ID; fails OOD semantics & Fails semantic navigation & Fails semantic navigation \\
\textbf{Q6: Sensor-dependent safety} & LiDAR gives the largest safety gain; strong LiDAR reliance & Depth improves safety; LiDAR adds no benefit & Fails across modality settings & Marginal full-modality gain; weak sensor integration & Small improvement from LiDAR; overall weak \\
\bottomrule
\end{tabularx}
\end{table*}

\begin{figure*}[!t]
\centering
\includegraphics[
  width=\textwidth,
  height=0.275\textheight
]{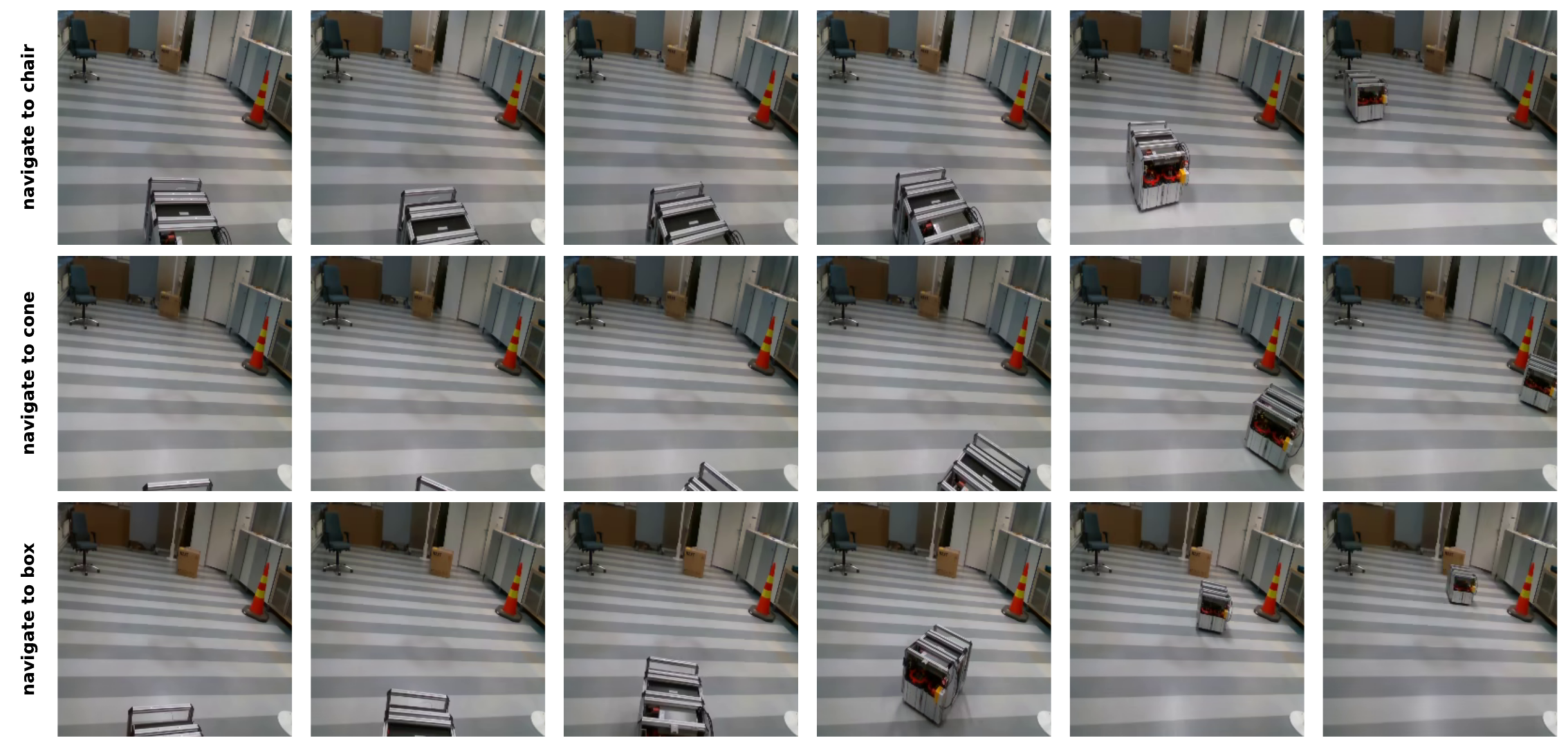}
\caption{Frame grid visualization of sim-to-real transfer on the Kelo Robile robot across three goal-navigation episodes.}
\label{fig:robile-frame-grid}
\end{figure*}

\textbf{Real-World Deployment}
We further validated the GR00T navigation policy on a physical mobile robot platform as a sim-to-real transfer policy, where the model is trained on simulated demonstrations and deployed directly on a real robot without environment-specific retraining. Figure~\ref{fig:robile-frame-grid} illustrates the results of three deployment episodes, each targeting a different goal location. Consistent with the camera-viewpoint evaluation, changes in camera viewpoint did not affect navigation performance in the real-world trials, provided that the target remains visible in at least one of the camera views and the robot's initial state stays within the training distribution.

\section{Conclusion}
\label{sec:conclusion}
In this paper, we introduced a unified benchmark for evaluating VLA models in mobile robot navigation across five task categories and controlled ID/OOD shifts. We compared GR00T N1.6, \(\pi\)0.5, OpenVLA-OFT, SmolVLA, and ACT as closed-loop language-conditioned navigation policies.
Our results show that strong ID performance does not imply broad OOD robustness. GR00T gives the best overall ID performance, handles many visual shifts well, and shows complex-language understanding, while \(\pi\)0.5 shows procedural flexibility and better spatial expansion beyond the ID training range. However, none of the models demonstrates reliable visual memory, and all models struggle with semantic target shifts and world-knowledge tasks. Modality-dropout results further show different sensing biases, with GR00T relying more on proprioceptive state and \(\pi\)0.5 more on visual observations.
Finally, real-world deployment with GR00T confirms that visual variations are tolerable when the target remains visible within the training distribution. Overall, VLAs can be adapted to mobile navigation, but robust deployment still requires stronger grounding across language, perception, state, and action.

\FloatBarrier

\bibliographystyle{IEEEtran}
\nocite{*}
\bibliography{references_crol}

\end{document}